# Rigel-Ladon: An Option-Count Law for a One-Pass Decision Encoder, Tested on Held-Out Option Counts and Held-Out Schemas

*Nested option sets, independence from irrelevant alternatives, position and rival effects, fixes that fail to transfer, and the twenty threshold tests of the first version*

Mohammed Nayeem Khan
mohammed.nayeem.khan@intel.com
mnknayeem@gmail.com

Vrishank Tiwari
vrishank.tiwari@intel.com

Navin Kumar R
navin.kumar.r@intel.com

Intel Corporation

**Abstract**

Laya is a small neural network that reads a text, a question and K written options, and returns one probability per option in a single pass. We ask how its accuracy falls as K grows from 2 to 35 and whether that can be predicted. We score nested option sets (every K-option question contains the smaller ones) on four question pools, for the released model (Base) and our fine-tune (FT), and score every item against each single rival. FT trained on one 12-option topic question (in-family); a 12-option HR (department-routing) question is out-of-family. Logit accuracy (the log of correct over incorrect answers) is linear in log(K-1) with slope 0.68 to 0.88. Fitted on K up to 6, the law predicts K=7 to 12 within 0.6 points and K up to 24 within 1.2 (a log-linear rule: 1.6 and 6.7). Taking the slope from the topic question and measuring only two-option accuracy on HR predicts its other K within 0.3 points (FT) and 0.7 (Base). On a 35-label pool of five schemas the law fails (5 to 11 points). Independence of irrelevant alternatives is rejected: the gold-versus-rival log-odds shifts by +0.24 to +2.07 nats per doubling of K, yet pairwise answers reproduce K-way accuracy within 4.7 points. Order averaging (averaging over several orderings of the options), shortlists, a temperature law and calibrated prediction sets did not transfer (FT's 90% sets cover 84% on HR). 2 of 10 predictions made in advance held. Caveats: one model family and LLM-labelled synthetic text.

**Key results (all values recomputed at build time)**

1. Logit accuracy is linear in log(K-1): slope c = 0.68 to 0.88 in all eight cells (intervals in Table 6 of Section 5). Fitted on K up to 6, it predicts K=7 to 12 within 0.6 points (log-linear rule 1.6, IIA with c=1 3.1) and K=16 to 24 on Pool-24 within 1.2 (6.7 for the log-linear rule).
2. With the slope borrowed from the topic schema and only two-option accuracy measured, the law predicts all other K on HR-12, a schema FT never saw, within 0.3 points (FT) and 0.7 (Base); a borrowed log-linear slope is off by 3.0 and 6.2. On Mega-35, fitted on two points, it fails by 5 to 11 points at K above 12.
3. IIA is rejected: the raw gap between the gold option and a fixed rival rises by +0.24 to +2.07 nats per doubling of K, with every interval excluding zero; composing the K-1 pairwise answers still reproduces K-way accuracy within 4.7 points and misses the probability by 9 to 23.
4. The strongest pairwise rival predicts K-way errors with AUROC 0.83 to 0.96, against 0.75 to 0.84 for the count of rivals that win; a rival from another schema is no easier than one from the same schema.
5. Position: the pre-stated spread test passes in 1 of 6 cells; an exploratory trend test shows earlier slots more accurate on HR-12 and Pool-24 by 6 to 8 points from first to last slot.
6. Fixes tested on held-out schemas: ten-order averaging gains at most 2.2 points with every interval including zero; shortlist-then-compare and a temperature law fitted on the topic schema do not help reliably; a 90% conformal set calibrated on topics covers 84% on HR-12 (FT) and 99% (Base, sets of 5.9 options).
7. 2 of 10 pre-stated predictions held (Section 10); the others are reported as refuted or half-confirmed.
8. Part II, the twenty threshold tests of the first version: released 5 pass / 8 fail / 1 inconclusive; fine-tuned 13 / 4 / 3. Every threshold is a convention, stated with its rationale next to the test.
9. The stored 11+ temperature (0.1006) is applied as 0.5; confidence at K=12 is 95% at 74% accuracy (T-02 fail); the in-sample ECE-optimal total temperature is about 2.2.
10. Short 24-option questions are insensitive to `head_max_len` (0 of 468 items change; head 185 tokens); long ones are at 5% (Base) to 4% (FT) at 192 and 38% for FT at 768, and the collapsed-options flag fires on 0% of the questions that lose every informative token.
11. The K=12 fine-tune gains +1.5 points on an unseen schema (p=0.541), short of the 5-point bar (T-14 fail); 443 in-schema rows give +6.6 (p=0.002); a TF-IDF baseline is within 0.7 points on HR.

# 1 Introduction

Accuracy on a multiple-choice question depends on how many options are listed, and the dependence is not the same for every model. Laya is a small encoder that reads a text together with a question and the written options in one forward pass and returns one probability for every option. The caller decides how many options to list, and the model card gives a temperature table for the option counts it was tuned on, up to 12. What an application needs to know before it lists 16 or 24 options is how much accuracy it will lose and whether the confidence can be trusted. The first version of this report answered with twenty pass-or-fail tests, each with a threshold that we had chosen, and the thresholds were the weakest part: a verdict of fail at 90% retention says little about how accuracy behaves at 13 options or at 35.

We keep those tests as a second part (Sections 11 to 21) and build the first part around quantities that can be predicted and checked. The experiments use nested option sets, so that a K-option question contains every smaller question of the same item (Section 4). We measured accuracy and the probability of every option from K=2 up to 35 on four pools of questions, two models and up to 10 draws per item, and we scored every item against every single rival.

We report five results.

1. Accuracy follows a law with one free exponent. Logit accuracy falls linearly in log(K-1), and the fitted slope is 0.68 to 0.88 in every pool and model. Fitted on K up to 6, the law predicts accuracy at K from 7 to 12 within 0.6 points on average, against 1.6 for a log-linear rule, and on Pool-24, the only pool of the four that goes past K=12 and stops at 24, within 1.2 points at K above 12, against 6.7. On the 35-label pool built from five schemas the law fails, by 5 to 11 points at K above 12 when it is fitted on two points of that pool (Section 5).
2. The exponent transfers. Taking it from the topic schema and measuring only two-option accuracy on a different schema (HR) predicts every other K there within 0.3 points for the fine-tune and 0.7 for the released checkpoint, where a borrowed log-linear slope is off by 3.0 and 6.2.
3. IIA fails in both models. The gap between the gold option and a fixed rival changes by +0.24 to +2.07 nats per doubling of K, with every interval excluding zero. The gap grows with K, so added options help the gold option against that rival, and the probability of one option has to be read together with the list around it (Section 6).
4. Position matters little by the test we pre-stated. The spread of accuracy over the slot of the gold option exceeds a permutation null in 1 of 6 cells. A trend test that we added afterwards finds that early slots are more accurate on HR-12 and Pool-24, by 6 to 8 points from the first to the last slot (Section 7).
5. The fixes behave differently on held-out schemas (Section 8). Several of our pre-stated predictions failed, and Section 10 lists which.

The question of what is in-family matters for every headline. The fine-tuned model (FT) trained on the topic schema, and Topic-12 is therefore its in-family test. HR-12 is out-of-family for FT, and Pool-24 and Mega-35 contain both kinds of item. The released checkpoint (Base) has no training data we can inspect. Whenever a number below comes from an in-family condition the text says so.

## 2 Related work and positioning

Searches covered Hugging Face Papers and arXiv at title and abstract level on 10 and 18 November 2026. Every paper cited below was opened, and what is said about it goes no further than its abstract, except for the five papers on option-set dependence, order dependence, per-option scoring, conformal sets and pairwise ranking, whose method and evaluation sections we also read. The search is not exhaustive.

### 2.1 Typed-decision models and their evaluations

A typed decision maps evidence, a question and caller-defined options to one probability per option in a single forward pass. Jev introduced the interface and Laya is an open encoder that implements it. Sharma et al. [2026] built TypedBench to score such models on accuracy across paraphrases, calibration against the noise floor of a perfectly calibrated predictor, and cost, and report that decoder baselines degrade as options are added. We borrow the noise-floor idea for the T-04 threshold. Rafe and Das [2026] find a fine-tuned DeBERTa-v3-large most accurate on almost every labelled benchmark they test; T-18 runs the same kind of check at twelve labels with a TF-IDF classifier.

### 2.2 Option order, position and option count

Selection bias and order sensitivity are established for decoder language models on multiple-choice questions [Zheng et al. 2023; Pezeshkpour and Hruschka 2023], and Ok et al. [2026] split an option-bias score into a flip rate and a positional skew. For typed decisions, Zhang et al. [2026] (AnyJev) report on two 20-option tasks that a readout from a pretrained language model favours some labels and some list positions. They correct both without training, by dividing out a label prior and averaging log-probabilities over the K cyclic rotations of the option list; the flip rate falls from 0.33 to 0.14 and 0.18, at the price of K prefills. Teke [2026] scores each option blind to the others, and a sequential head trained with shuffled options still changed 2.3 to 2.8% of answers when only the order changed. Li and Wang [2026] (Jeva) require permutation invariance and candidate sets that scale to thousands of options.

T-11 and T-12 measure the same two defects on the sequential head of Laya at K=12. The released checkpoint flips 31% of items over 24 random orderings, a shuffle-trained fine-tune 11%, and the parallel layout 0%. The released checkpoint's share is close in size to the 0.33 that AnyJev reports for unmodified readouts, but the models, the tasks and the flip statistic differ, so the numbers are not comparable. Section 8 runs prior division and averaging over random orderings on the nested draws. A majority vote over the 24 orderings of T-11, a rough analogue of rotation averaging, moves accuracy by +1.4 points for Base and -0.1 for FT (Table 24).

The effect of the option count itself is less settled. MMLU-Pro [Wang et al. 2024] raised the choice set from four to ten options together with harder questions, and accuracy fell by 16 to 33%, so the two changes are confounded; Song et al. [2025] report that the number of options influences accuracy in a study of five language models; Gao et al. [2026] refine ordinal scales from K=2 to 14 with items and source scores held fixed and find that scale utilisation falls for every model they test. The K-curves in Section 14 also hold the item fixed and change only the sampled distractors, for a nominal question and an encoder, from K=2 to 12 and (with a second schema) to 24, and they report chance-adjusted skill.

### 2.3 Labels, descriptions and the token budget

Azizi et al. [2026] show that the probability in Jev-style typed-decision models follows the option labels more than their written definitions, and they trace the effect in Laya to the rendering of each option as a label followed by its definition. T-07 builds the converse case: the names carry no information, so the description is the only evidence, and the head budget cuts the description. The controls in Table 26 (names only against descriptions only) point the same way for the released checkpoints. The backbone, ModernBERT, was trained with a native sequence length of 8192 [Warner et al. 2024], so the default 192-token head and 512-token sequence are runtime settings. Liu et al. [2023] show that decoder models use long contexts less well as they grow, which is a possible but untested reading of our loss at a 1536-token head. Milios et al. [2023] handle a large label space in an in-context classifier by showing only a retrieved part of it, the same idea as the shortlist helper that the Laya runtime ships; Section 8 tests a shortlist built from the encoder's own scores.

### 2.4 Calibration

Temperature scaling is the standard post-hoc fix for over-confident networks [Guo et al. 2017]; Kull et al. [2019] propose a natively multiclass alternative to a single scalar, and Desai and Durrett [2020] analyse the calibration of pre-trained transformers in and out of domain. Teke [2026] finds that temperature scaling lowers calibration error on a development set but raises it on held-out questions. T-04 fits its multiplier on the same 270 items that it scores, so the optimum is optimistic. It is used only to show the direction and size of the error of the shipped scale. Errors in test labels can reorder systems [Northcutt et al. 2021], the reason for T-19.

### 2.5 Independence of irrelevant alternatives, shortlists and conformal sets

Rajput et al. [2026] test independence from irrelevant alternatives (IIA) on a hosted Jev model with 299,463 calls on intent and topic classification, and find that every added option moves the log-odds between the gold label and its strongest wrong rival, with a distortion that grows with the logarithm of the number of options and reaches about 1 nat at 255 options. Huang et al. [2026] show that the order dependence of System One models grows with the number of options and that cyclic averaging reduces it without removing it; their Runoff method scores every option alone to build a shortlist and compares the shortlisted options in listings where each takes every position. Stekel [2026] compares joint and per-option scoring in compact encoders and reports that joint encoders change 12 to 23% of their answers when the options are reordered. Qin et al. [2023] use pairwise comparisons to rank candidates with decoder models, which is the decomposition that our composition test applies to a choice question. Kumar et al. [2023] apply conformal prediction to multiple-choice questions and examine out-of-subject questions, where exchangeability can fail.

Section 6 repeats the IIA test on an open encoder with a nested design and adds a composition test that turns pairwise results into a K-way prediction. Section 5 fits a law of accuracy against K and tests it on K and schemas held out of the fit. Section 8 runs the order averaging, shortlist and conformal procedures of this literature on the same questions.

### 2.6 Position of this report

The IIA question and the order dependence are earlier findings; our contribution is to test them on an open model with a design that holds the question fixed while K changes, and to turn the K-dependence into a fitted and checked law. Within the searches above we found no earlier work on the token budget that cuts the option block of a typed-decision encoder, on the stored and applied option-count temperature of Laya, or on schema transfer of a Laya fine-tune; those measurements, the fitted exponent and its transfer, and the held-out tests of the fixes are what we add. The measurements are one encoder family on synthetic, LLM-labelled data, so the exponent is a result about Laya on these schemas and not a constant of the field.

## 3 System under test, data and conventions

### 3.1 Component under test

Laya reads a question and a piece of text together. The question is rendered as [CLS] choice question: <instruction> [SEP] [MASK] name: description [MASK] name: description … [SEP] <text> [SEP]. A ModernBERT-large encoder (28 layers) processes the whole sequence once. The hidden state at each [MASK] marker is the state of that option; a two-layer transformer head lets the option states attend to one another and a scorer emits one logit per option. Logits are divided by a temperature that depends on the question type and on how many options the question has (buckets 2, 3-5, 6-10, 11+), then soft-maxed; the top probability is reported as answer_confidence. Because every option is read from a position inside one bounded sequence, two budgets matter once K grows: how many tokens the option block may occupy and how many the whole sequence may.

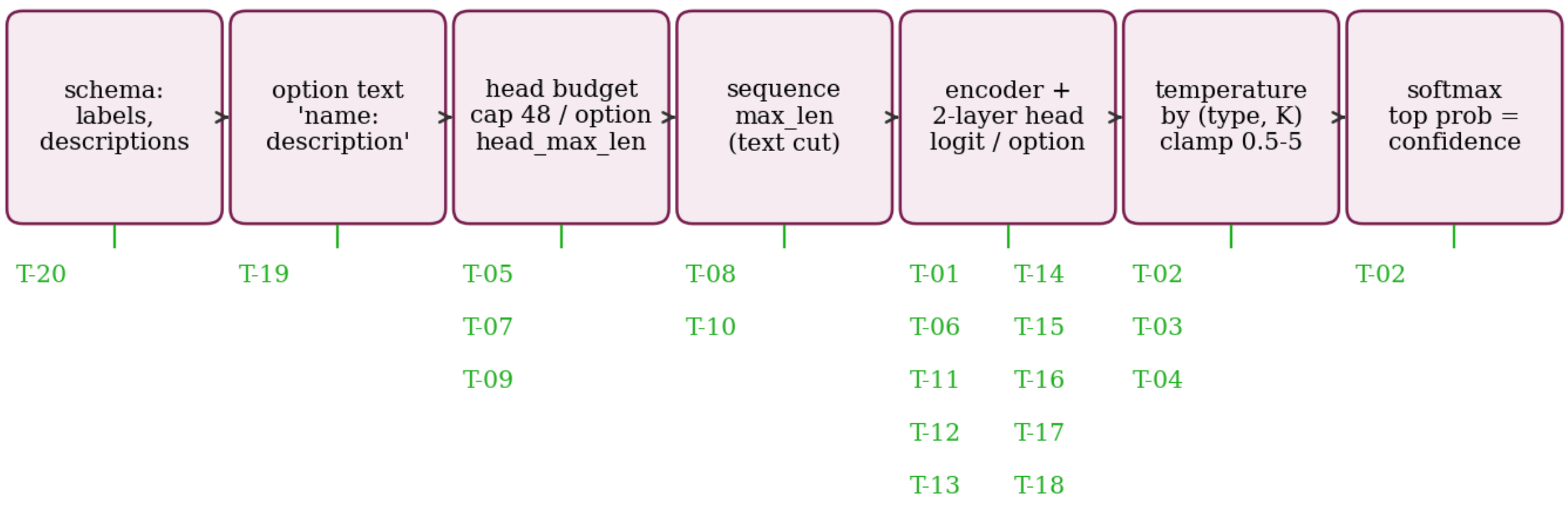

**Figure 1.** Test rig: the path of a K-option question through the runtime, left to right, and the test cases (red) that probe each stage. T-03 and T-10 read configuration and call the runtime's own functions without a model; all other tests replay frozen forward passes.

**Table 1.** Runtime knobs relevant to many-option questions (released root checkpoint, laya 0.4.26).

| Knob | Value | What it controls |
|---|---|---|
| head_max_len | 192 | tokens shared by the instruction and all options; options are cut once their total exceeds head_max_len−16 |
| per-option cap | 48 tokens | applied first; fixed in build_head |

| Knob | Value | What it controls |
|---|---|---|
| floor | 4 tokens | no option is cut below 4 tokens (`max(4, (head_max_len-16)//K)`) |
| `max_len` | 512 | total sequence; the state is cut from the right to fit what the head leaves |
| `temperature_by_options` | 6 buckets | per (type, K) divisor on the logits; clamped to [0.5, 5.0] before use |
| `usage["options"]` | report | non-empty only when two options become token-identical after cutting |

### 3.2 Systems and data

**Table 2.** Systems under test and the labels used in this report.

| Label | What it is |
|---|---|
| **Base** | released root checkpoint (`convaiinnovations/laya`) |
| **TD** | released `typed-decisions` sub-checkpoint |
| **ML** | released `multilingual` sub-checkpoint |
| **FT** | our fine-tune of the root checkpoint with Laya's trainer (2,818 rows, 4 epochs) |
| **FT+HR** | FT plus 443 rows of the 12-label HR schema (4 epochs) |
| **FT-noshuf** | FT trained without option shuffling (ablation) |
| **FT-par** | FT trained with the parallel option layout (architectural variant) |
| FT seeds 1-3 | FT retrained with trainer seeds 1, 2, 3 (same data) |

The test material has two parts. The *topic schema* is a 12-way support-ticket question whose option descriptions are the labels themselves, so options are 3-5 tokens long; the *HR schema* is a 12-way question with keyword descriptions (7-11 tokens). Merged, they form a 24-label pool used for K=16 and K=24. Qwen3.8-27B produced the texts and labels. For each label the teacher writes ten messages that should carry it, a second blind call labels the same messages without seeing the intended label, and every message on which the two calls disagree is dropped. The held-out sets never enter training, and the training texts were de-duplicated against them. The most similar training row to a topic test item has median character-5-gram Jaccard 0.12, and 1% of test items reach 0.5 or more. Training used Laya's own trainer (RLCD loss, 4 epochs, micro-batch 8 with 8 accumulation steps, encoder learning rate 2.5e-05, head learning rate 0.0001) on 2,818 rows, 1145 of them from the topic schema. The FT+HR model adds the 443 HR rows. One RTX 4090 trains a model in about five minutes.

**Table 3.** The two 12-label schemas. Topic: label equals description. HR: each label carries a short keyword description.

| Schema | Labels (K=12) | Test items | Train rows | Teacher yield (kept / generated) |
|---|---|---|---|---|
| Topic (support tickets) | login, password, shipping, returns, invoice, refund, api, latency, security, privacy, feature_request, praise | 195 | 1145 | 195 / 240 (81%) |
| HR requests | payroll (salary, pay slips, deductions); benefits (insurance, retirement, perks); leave (vacation, sick leave, parental leave); recruiting (open roles, referrals, interviews); onboarding (new hire setup, orientation); offboarding (resignation, exit, final settlement); training (courses, certifications, learning budget); performance_review (ratings, feedback, promotion cycles); workplace_safety (injuries, hazards, safety rules); expenses (reimbursements, travel claims); it_access (accounts, laptops, passwords, permissions); policy_question (general HR rules and compliance) | 273 | 443 | 716 / 773 (93%) |

### 3.3 Statistics and verdict rules

Proportions carry Wilson 95% intervals. Where several draws come from one item (K-curves use three random option subsets per item) the interval is a bootstrap over items, not over draws. Paired comparisons on identical items use the exact McNemar test. Calibration is measured by expected calibration error (ECE, ten equal-width bins) on the top probability. Each test has a threshold and one of three verdicts, and the rule that turns an observed value into a verdict depends on the kind of test.

- Interval tests (T-01, T-02, T-04, T-07, T-08, T-09, T-11, T-19): **pass** if the whole 95% interval is on the safe side of the threshold, **fail** if the point estimate is on the wrong side, **inconclusive** in between.
- Band tests (T-06, T-16): pass if the interval lies inside ± the band, fail if the point estimate lies outside it. T-12 and T-17 have no interval; their inconclusive zones are 15% around the limit (T-12) and sd from 2 to 3 points (T-17).
- Effect tests (T-13, T-14, T-15, T-18): pass if the paired gain is at least five points and $p<0.01$, fail if neither holds or the gain is not positive, inconclusive if one holds.
- Deterministic and control tests (T-03, T-05, T-10, T-20): T-05 passes only with zero changed items. T-10 passes when the runtime's own functions drop markers. T-20 passes when chance lies inside the interval. T-03 passes only when the applied value equals the stored one and is not on a clamp bound.

The thresholds are conventions fixed when the test plan was written, and they are the same for every system. The matrix states the rationale of each one next to the test, and Section 12.1 shows how far each can move before its verdict changes. A verdict says on which side of a stated line an observed value falls, or that the data cannot tell; it is not a general judgement of the model.

## 4 Design of the option-count experiments

The first report measured each K with its own random draw of distractors. That makes consecutive K values hard to compare, so we replaced the sampler. For every item and every draw we shuffle the distractors once and take the first K-1 of them together with the gold option. The K-option question therefore contains the (K-1)-option question, and one item at K=12 can be read as the 11 smaller questions it contains. The options are listed in one fixed random order per draw, and the gold option sits in a uniformly random slot, so the position effects of Section 7 are measured on the same questions. All K questions of an item are scored in one batched call.

We use four pools (Table 4). Topic-12 is the 12-label topic question used in the first report; the fine-tuned model trained on it, so it is the in-family condition. HR-12 is a 12-label department-routing question that the fine-tune never saw, and it is our out-of-family condition. Pool-24 and Mega-35 merge label sets of several schemas, so that K can pass anything seen in training. Each K-curve in this report comes from the stored probabilities of every option, not only the top answer: 945 distinct items, used in one to four of the pools, between 3 and 10 draws each. We also scored every item against every single rival as a two-option question (Section 6), 11 or more questions per item.

**Table 4.** The four option pools. K is the number of listed options; a draw is one random ordering of the distractors. In-family means that the fine-tuned model (FT) saw the same schema in training; the released checkpoint (Base) has no stated training data we can inspect.

| Pool | Labels | Items | K values | Draws | In-family for FT? |
|---|---|---|---|---|---|
| **Topic-12** | 12 | 270 | 2 to 12 (every K) | 10 | yes (schema seen in training at K=12) |
| **HR-12** | 12 | 273 | 2 to 12 (every K) | 10 | no (department routing, never trained on) |
| **Pool-24** | 24 | 468 | 2, 3, 4, 6, 8, 12, 16, 20, 24 | 5 | half topic items, half HR items; K above 12 was never trained |
| **Mega-35** | 35 | 870 | 2, 4, 8, 12, 16, 24, 35 | 3 | topic, HW and JSON items are in-family; HR, sentiment and mixed-language items are not |

Before running the new experiments we wrote ten predictions with a test for each (Table 5). Several came from outside this literature: independence from irrelevant alternatives (choice theory), the statistics of maxima, conformal prediction and permutation averaging. We report the ones that fail as well as the ones that hold. The predictions were fixed before the pairwise, shortlist, HR, Pool-24 and Mega-35 runs, but after the first report had shown the shape of the topic curve, so P3 is not a blind prediction for the topic schema.

**Table 5.** Predictions written down before the runs (`reviews/a6_ideas.md`); the verdict column is filled in by the scorecard in Section 10.

| ID | Claim | Pre-stated test |
|---|---|---|
| P1 | IIA: the K-way probability of the gold option is the composition of its K-1 pairwise odds | composed K=12 accuracy within 3 points for FT, more than 3 points off for Base |
| P2 | gold-versus-rival log-odds is constant in K (IIA) | slope per doubling of K is zero; its interval excludes 0 |
| P3 | logit accuracy is linear in log(K-1); fitted on K up to 6 it predicts K=8 to 12 and K up to 24 | held-out error under 2 points for K up to 12, and larger for K far beyond the fit |
| P4 | the slot of the gold option changes accuracy | the spread over slots exceeds a permutation null (Base more than FT) |
| P5 | one strong rival, not the number of rivals, causes the errors | the strongest pairwise rival predicts K=12 errors better than the count of rivals |
| P6 | averaging over option orders helps | at least 2 points for Base at 10 orders |
| P7 | shortlist then compare helps | at least 3 points for Base at K=12 |
| P8 | a temperature law fitted on one schema transfers | lower ECE than the shipped scale on the other schema |
| P9 | split-conformal sets calibrated on one schema keep their coverage on another | coverage within 2 points of the target on HR |
| P10 | a larger head budget repairs the 35-option pool | within 3 points of the K=24 curve |

## 5 An option-count law

The simplest model of a K-way choice says that the gold option beats K-1 rivals that all look alike. If the option scores are the logits of one softmax, the odds of the gold option are the gold-versus-rival odds divided by K-1, so that logit accuracy falls with log(K-1) with slope exactly -1. Real rivals are not alike, and the confusable ones matter more than the rest, so we expect the effective number of rivals to grow more slowly than K-1. We therefore fit logit(accuracy) = a - c log(K-1) with c free. An IIA scorer with exchangeable rivals would give c = 1, and a model that ignores the extra options would give c near 0.

Figure 2 shows the curves and Figure 3 the fit. The fitted exponent sits between 0.68 and 0.88 in all 8 cells of the nested runs (two models, four pools), although the cells differ in schema, item count and chance level; its interval excludes 1 in 5 of them (Table 6), so for this encoder rivals are not exchangeable. Fitted on K up to 6, the law predicts K from 7 to 12 with a mean error of 0.6 points over the 6 cells outside Mega-35 that have such K, against 1.6 for the log-linear rule and 3.1 for the IIA law with c = 1. On Pool-24 the same fit predicts K=16 to 24, which no fit point is near, with a mean error of 1.2 points, against 6.7 for the log-linear rule, 10.2 for a straight line in K and 7.3 with c = 1 (Figure 4). On Mega-35, where the fit has only two points (K=2 and 4) and the labels come from five schemas, the error is 2.9 points at K=8 and 12 and 8.2 at K above 12: the law is optimistic at K=35 for both models. P3 is therefore supported up to K=24 and refuted at K=35 on the mixed pool.

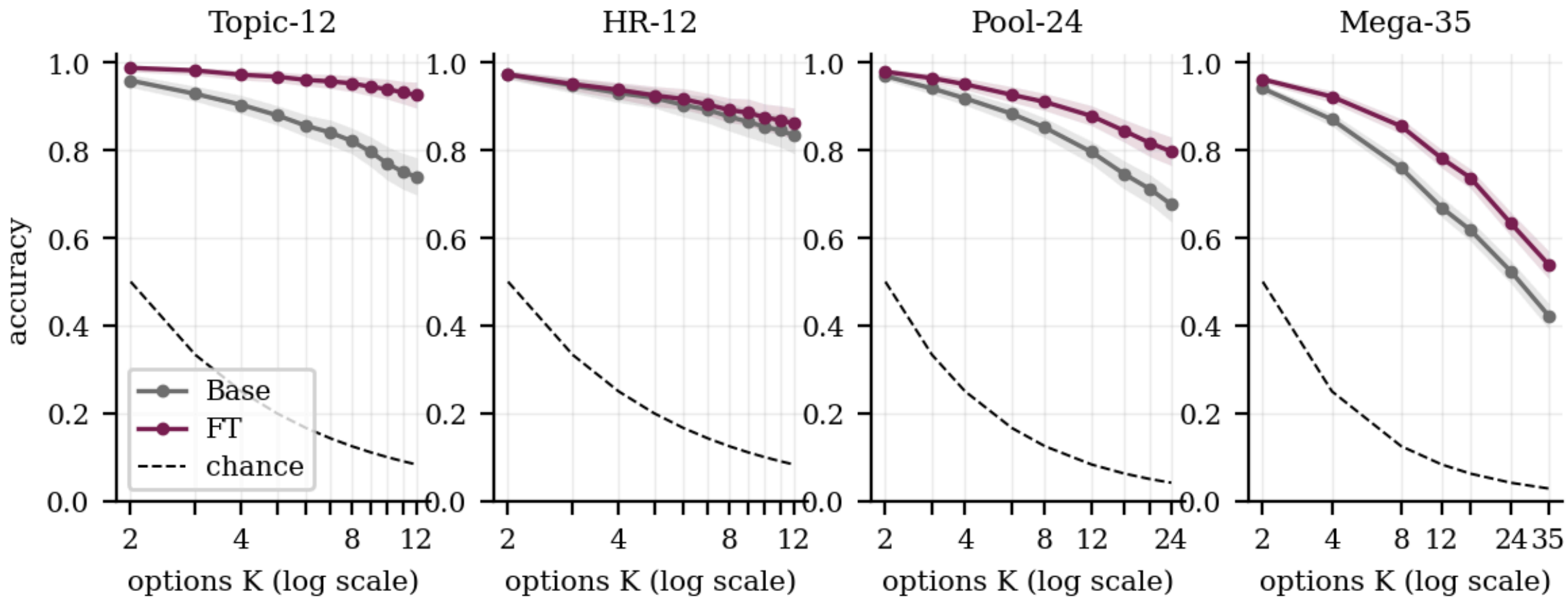

**Figure 2.** Accuracy against K in the four pools (95% bootstrap bands over items, all draws pooled). The dashed line is chance, 1/K. Topic-12 is in-family for FT; the other three pools are not, or only in part.

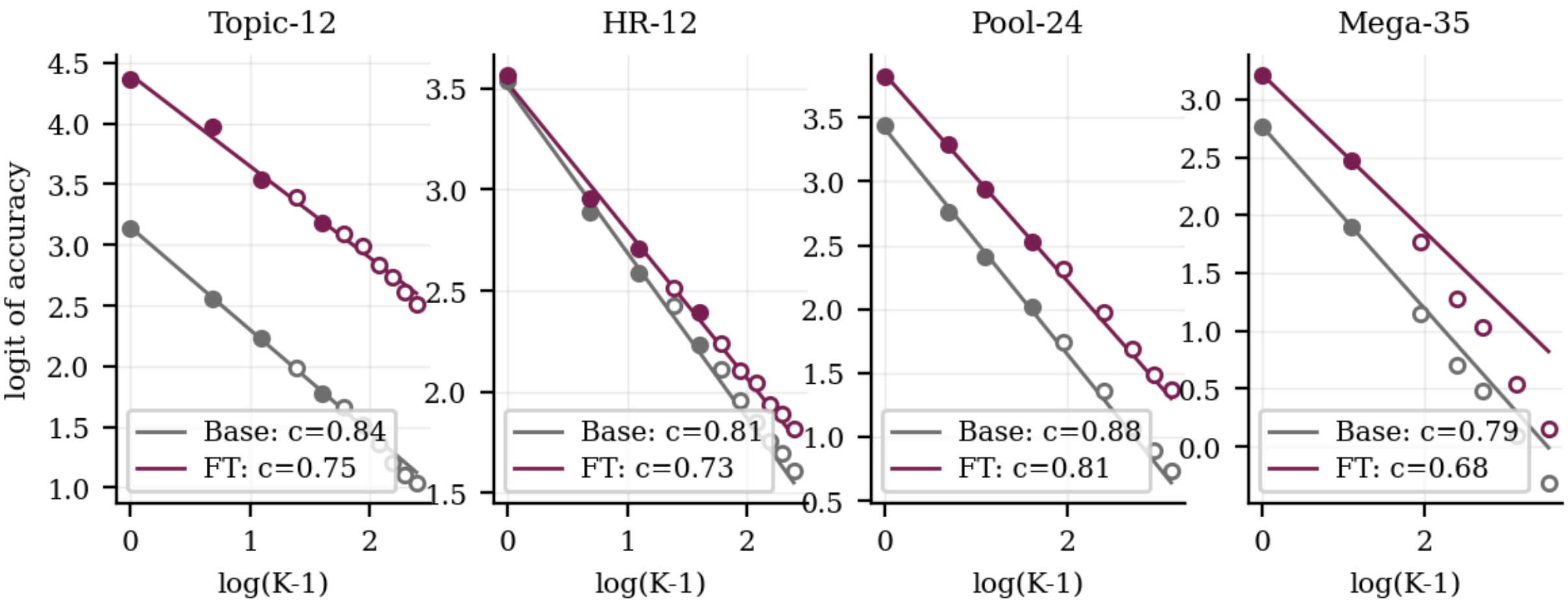

**Figure 3.** The option-count law. Logit accuracy against log(K-1) in each pool; filled dots are the K values used for the fit (K up to 6), open dots are held out. The line is the fit and its extrapolation; c is the fitted slope magnitude.

**Table 6.** Fitted exponent c and held-out error of five laws fitted on K up to 6 only. Each law cell gives the mean absolute error in accuracy points on K from 7 to 12 (near) and on K above 12 (far); n/a means there is no such K in the pool. c has a 95% bootstrap interval over items.

| Pool, model | c | IIA, c fixed at 1 | log-odds law, c free | log-linear in K | linear in K | power law of the error |
|---|---|---|---|---|---|---|
| Topic-12 Base | 0.84 [0.67 to 1.03] | 2.4 / n/a | 0.9 / n/a | 2.9 / n/a | 1.8 / n/a | 0.4 / n/a |
| Topic-12 FT | 0.75 [0.46 to 1.16] | 1.7 / n/a | 0.2 / n/a | 0.7 / n/a | 0.6 / n/a | 0.2 / n/a |
| HR-12 Base | 0.81 [0.66 to 0.97] | 3.5 / n/a | 0.6 / n/a | 1.4 / n/a | 1.7 / n/a | 1.1 / n/a |
| HR-12 FT | 0.73 [0.57 to 0.90] | 4.0 / n/a | 0.2 / n/a | 1.2 / n/a | 1.3 / n/a | 0.5 / n/a |
| Pool-24 Base | 0.88 [0.75 to 1.04] | 3.4 / 7.1 | 0.9 / 1.6 | 2.3 / 8.1 | 2.8 / 12.6 | 2.0 / 7.5 |
| Pool-24 FT | 0.81 [0.66 to 1.00] | 3.5 / 7.6 | 0.6 / 0.7 | 1.1 / 5.3 | 2.1 / 7.7 | 1.0 / 2.6 |
| Mega-35 Base | 0.79 [0.65 to 0.93] | 4.6 / 8.0 | 2.6 / 5.4 | 6.4 / 16.6 | 5.7 / 31.9 | 0.2 / 9.1 |

| Pool, model | c | IIA, c fixed at 1 | log-odds law, c free | log-linear in K | linear in K | power law of the error |
|---|---|---|---|---|---|---|
| Mega-35 FT | 0.68 [0.52 to 0.85] | 4.9 / 7.2 | 3.2 / 11.1 | 5.3 / 18.4 | 1.4 / 12.9 | 2.5 / 7.2 |

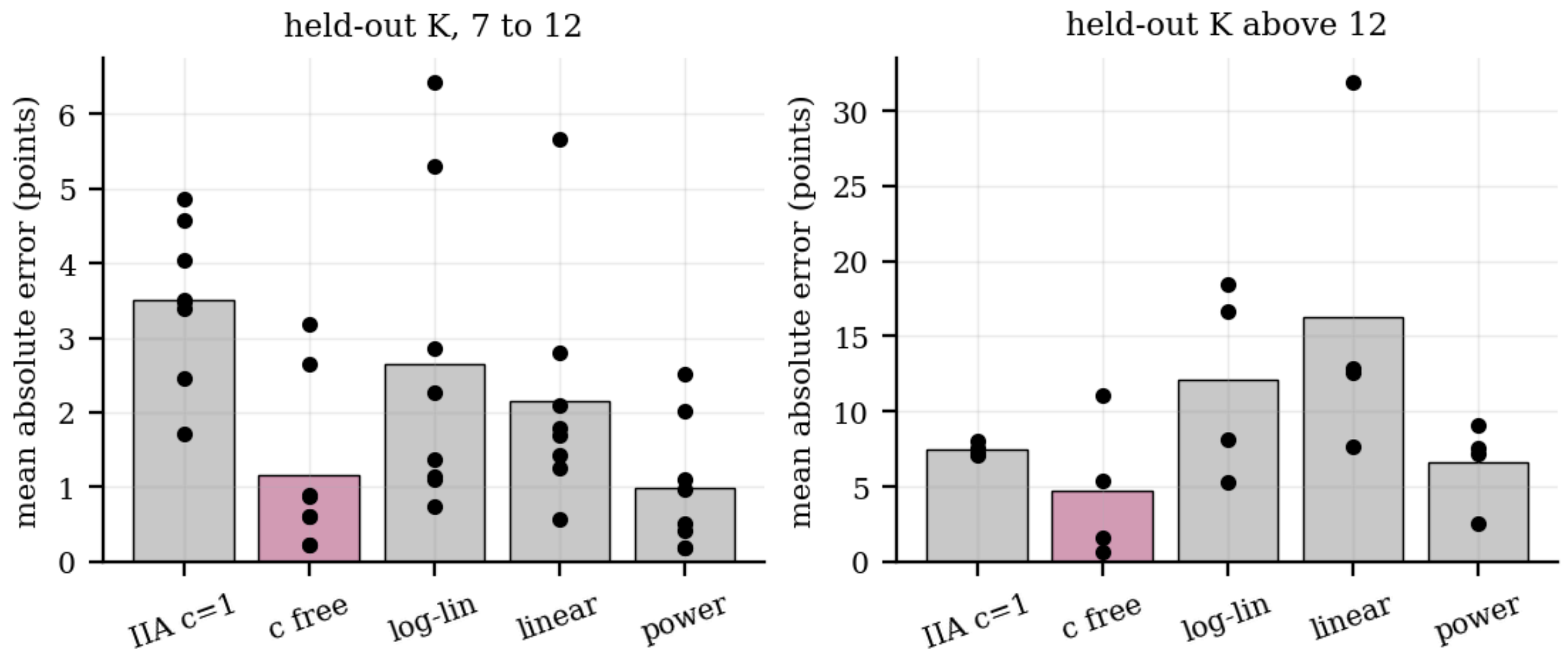

**Figure 4.** Held-out error of five laws fitted on K up to 6. Bars are means over the pool and model cells; dots are the individual cells (up to eight). The log-odds law with a free exponent is highlighted.

A second check asks whether the exponent belongs to the model or to the schema. We take c from the topic curve of each model, measure only the two-option accuracy of the target pool, and predict every other K there. On HR-12, which FT never saw, the mean error is 0.3 points for FT and 0.7 for Base, against 3.0 and 6.2 for a log-linear slope borrowed the same way and 5.2 and 4.1 for c = 1 (Figure 5). On Pool-24 the borrowed exponent errs by 0.8 (FT) and 0.9 (Base) points over K from 3 to 24, and on Mega-35 by 3.8 and 0.9, so the exponent taken from a six-point curve beats the one fitted on two points there. This gives a budget rule that needs one measurement on a new schema: take the two-option accuracy, use c of about 0.75, and read off the K at which accuracy crosses your floor. The rule is a fit to these schemas, not a derived constant. At K=35 on Mega-35 it predicts 64% for FT and 45% for Base against 54% and 42% measured.

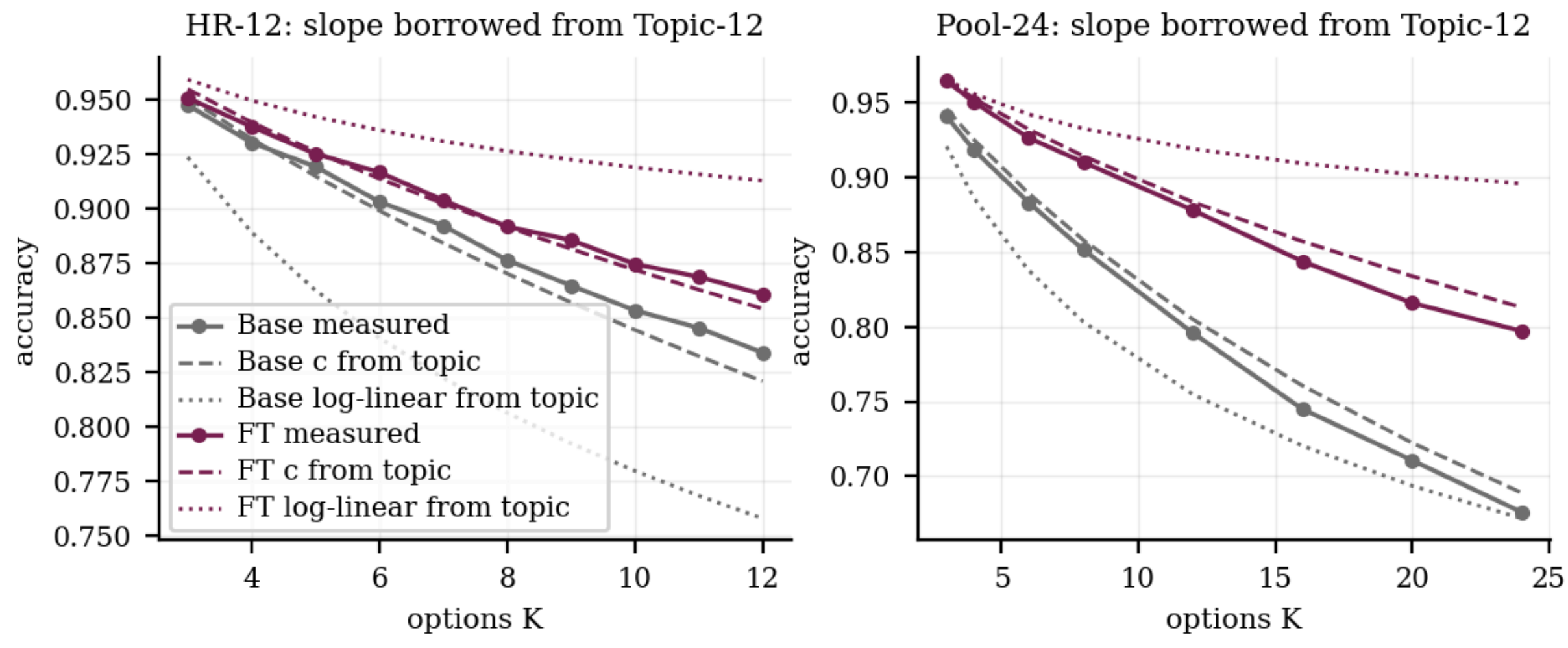

**Figure 5.** Transfer to a held-out schema. The exponent c (or the log-linear slope) is taken from the topic curves; the only quantity measured on the target pool is its two-option accuracy. Dashed: log-odds law; dotted: log-linear rule; solid: the measurement.

Two further checks use data that the first report already held. Reversing the direction, with the exponent taken from HR-12 and the two-option accuracy measured on the target, the mean error is 1.1 (Base), 0.2 (FT) points on Topic-12 and 1.8 (Base), 1.4 (FT) on Pool-24, against 3.7 (Base), 4.2 (FT) and 3.9 (Base), 1.5 (FT) for the borrowed log-linear slope. Across the 8 systems of Table 7 (the two released sub-checkpoints, the multilingual and typed-decisions ones, three more FT seeds and FT+HR), the exponent lies between 0.61 and 0.99 on the two schemas, wider than in the nested data and with fewer points per curve. Held-out K=8 and 12 are predicted with a mean error of 1.5 points against 1.9 for the log-linear rule, and the law is the better of the two in 11 of 16 cells. The advantage of the law over the log-linear rule is larger beyond K=12 and when the exponent is borrowed from another schema than at K=8 and 12 alone. These curves reuse the draws of the first report and are not independent of the nested data for the same systems, but they cover six systems that the nested runs do not.

**Table 7.** The law on the other released and fine-tuned checkpoints, from the K-curves of the first report (K up to 6 fitted, K=8 and 12 held out; three random draws per item, so these curves are not the nested draws). FT s1 to s3 are fine-tunes with other training seeds; FT+HR also saw HR rows in training, so its HR column is in-family.

| System | c (topic) | Error K=8, 12 (law) | Error (log-linear) | c (HR) | Error K=8, 12 (law) | Error (log-linear) |
|---|---|---|---|---|---|---|
| **Base** | 0.83 | 1.1 | 2.2 | 0.89 | 1.4 | 2.4 |
| **TD** | 0.69 | 2.4 | 3.4 | 0.84 | 1.2 | 4.0 |
| **ML** | 0.96 | 3.0 | 1.0 | 0.61 | 3.5 | 4.3 |
| **FT** | 0.82 | 1.4 | 1.2 | 0.96 | 3.0 | 0.8 |
| **FT+HR** | 0.92 | 0.7 | 0.6 | 0.73 | 0.1 | 1.5 |
| **FT seed 1** | 0.86 | 0.1 | 0.9 | 0.68 | 0.6 | 2.2 |
| **FT seed 2** | 0.89 | 0.3 | 0.6 | 0.91 | 0.9 | 3.0 |
| **FT seed 3** | 0.91 | 0.4 | 1.2 | 0.99 | 3.2 | 1.5 |

## 6 Independence from irrelevant alternatives and the rivals that matter

IIA says that adding an option which is neither the gold answer nor its strongest rival leaves the odds between those two unchanged. Rajput et al. [2026] ran this test on a hosted decision model with 299,463 calls and found that every added option moves the gold-versus-rival log-odds, with a shift that grows with the logarithm of the number of options. We repeat the test on an open encoder with the nested design: the rival is fixed at the first distractor of each draw, and the other K-2 options are added one at a time. The runtime divides the logits by a temperature that depends on K for Base (1.91 at K=2, 1.76 for 3 to 5, 1.00 for 6 to 10 and 0.5 above), so we convert every probability back to its raw logit gap before comparing K values. FT uses one constant (5.0).

IIA does not hold for either model (Table 8). The gap between gold and rival grows with K, by +0.80 to +2.07 nats per doubling for FT and +0.24 to +0.54 for Base, and no interval includes zero. Rajput et al. report the same direction for an option that resembles the wrong answer: it draws probability away from that answer and toward the gold label. We did not control the similarity of the added options, so we cannot say that the same mechanism is at work. P2 predicted a negative slope for Base and a smaller slope for FT than for Base. Both parts were wrong: the rejection of IIA holds, but the sign is positive and FT moves more than Base. A positive slope points the same way as an exponent below 1 in the law of the previous section: accuracy then falls more slowly than it would if every added option competed equally with the gold option.

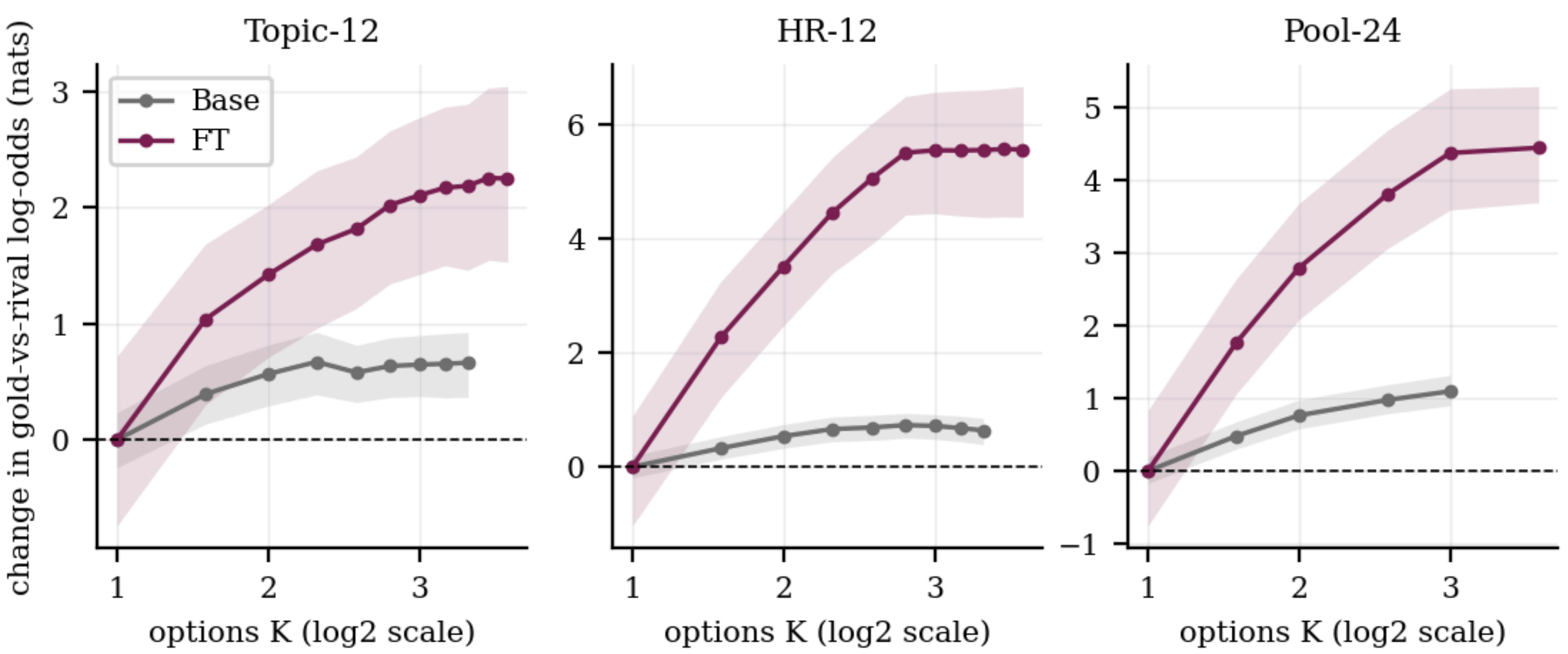

**Figure 6.** Does the gold-versus-rival log-odds stay put as options are added? Mean change in raw logit gap (probability log-odds multiplied by the temperature that the runtime divided by) from K=2, for the rival that is the first distractor of each draw. Bands are 95% bootstrap intervals over items. Base stops at K=10 because its stored temperature changes at K=11, and probabilities below 1e-4 were not stored.

**Table 8.** Slope of the gold-versus-rival log-odds per doubling of K (nats, raw logit scale) and its endpoints. IIA predicts a slope of 0.

| Pool, model | Slope per doubling [95% CI] | Gap at K=2 | Gap at largest K | Largest K |
|---|---|---|---|---|
| Topic-12 Base | +0.24 [0.18 to 0.32] | 5.3 | 6.0 | 10 |
| Topic-12 FT | +0.80 [0.58 to 1.02] | 25.9 | 28.2 | 12 |
| HR-12 Base | +0.28 [0.21 to 0.33] | 3.6 | 4.3 | 10 |
| HR-12 FT | +2.07 [1.80 to 2.34] | 16.7 | 22.3 | 12 |
| Pool-24 Base | +0.54 [0.48 to 0.61] | 4.3 | 5.4 | 8 |
| Pool-24 FT | +1.75 [1.55 to 1.99] | 20.1 | 24.5 | 12 |

The composition test asks how far IIA is from the truth in accuracy. We scored every item against each of its rivals separately, and for each K-option set composed those K-1 pairwise odds into a K-way probability (raw logit gaps summed in the softmax). A set counts as correct when the gold option wins every pairwise question (Figure 7, Table 9). At the largest K the composed accuracy is within 0.7 to 4.7 points of the measured one, and it is lower than the measured accuracy in 5 of 6 cells. It is a poor model of the probability: the composed probability of the gold option misses the measured one by 9 to 23 points on average. P1 predicted that FT would be within 3 points and Base beyond 3. Base is within 3 points in 3 of 3 cells and FT is beyond 3 in 2 of 3, so the prediction failed. The answer of the K-way question is close to what the pairs imply and the probability attached to it is not, so a user who reads one probability needs the list it was computed over.

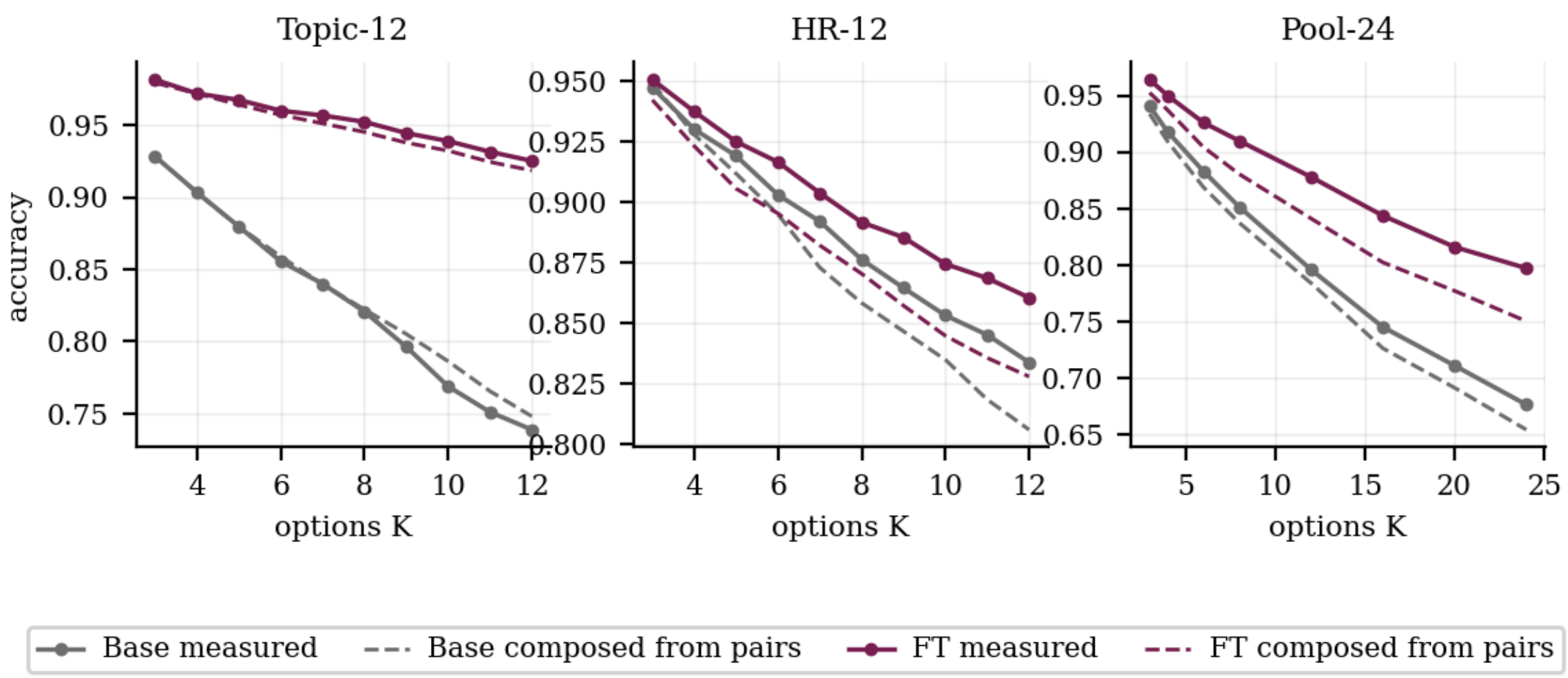

**Figure 7.** Composition test. Solid: measured accuracy of the K-way question. Dashed: the share of (item, draw) sets in which the gold option beats every one of its K-1 rivals when asked as a two-option question, which is what IIA predicts the K-way answer to be.

**Table 9.** Composition at K=12 for the 12-label pools and K=24 for Pool-24. Composed accuracy is the share of sets in which gold wins every pairwise question; the gap is composed minus measured with a 95% interval. The next column is the mean absolute error between the composed and the measured probability of the gold option (points), and the last the share of sets on which the two accuracy indicators agree.

| Pool, model | K | Measured acc. | Composed acc. | Gap [95% CI] | \|P(gold) error\| | Agree (%) |
|---|---|---|---|---|---|---|
| Topic-12 Base | 12 | 73.9 | 74.8 | +0.9 [-2.9 to 4.9] | 15.6 | 84 |
| Topic-12 FT | 12 | 92.5 | 91.9 | -0.7 [-3.5 to 1.6] | 9.2 | 94 |
| HR-12 Base | 12 | 83.4 | 80.6 | -2.8 [-6.2 to 0.8] | 15.4 | 87 |
| HR-12 FT | 12 | 86.0 | 82.8 | -3.3 [-6.5 to -0.3] | 22.0 | 91 |
| Pool-24 Base | 24 | 67.6 | 65.4 | -2.2 [-5.7 to 1.3] | 23.0 | 78 |
| Pool-24 FT | 24 | 79.7 | 75.0 | -4.7 [-8.0 to -1.8] | 22.2 | 86 |

The errors follow the rival that wins. At the largest K the strongest pairwise rival predicts the K-way error with an AUROC of 0.83 to 0.96, against 0.75 to 0.84 for the number of rivals that win a pairwise question (Figure 8). The pairwise view does not explain every error: in 22 to 38% of the K-way errors the gold option wins every pairwise question, so the loss came from the options acting together. The confusable pairs are somewhat concentrated: the top 10% of ordered label pairs hold 32% (Base) and 43% (FT) of the pairwise losses on the topic schema, and 18% and 21% on HR (Figure 9). The schema of the rival does not matter: on Pool-24 the gold option wins 97.4% of the pairs against a rival from its own schema and 97.6% against one from the other schema for FT (96.7% and 96.9% for Base). The intervals overlap, so a rival from a different domain is no easier than one from the same domain.

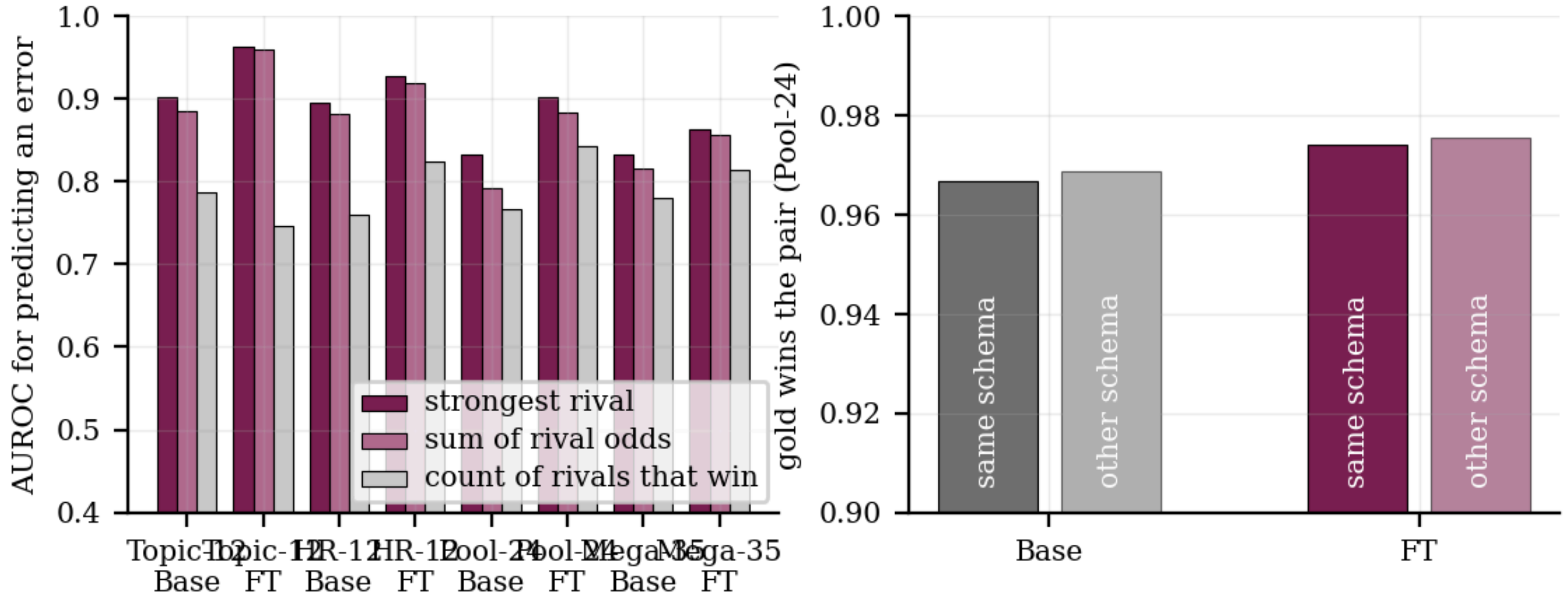

**Figure 8.** Which rival causes the errors. Left: AUROC of three scores of the pairwise questions for predicting that the K-way answer is wrong (K=12 for topic and HR, K=24 for Pool-24 and Mega-35). Right: on Pool-24, the share of two-option questions the gold option wins when the rival comes from the gold option's own schema or from the other one.

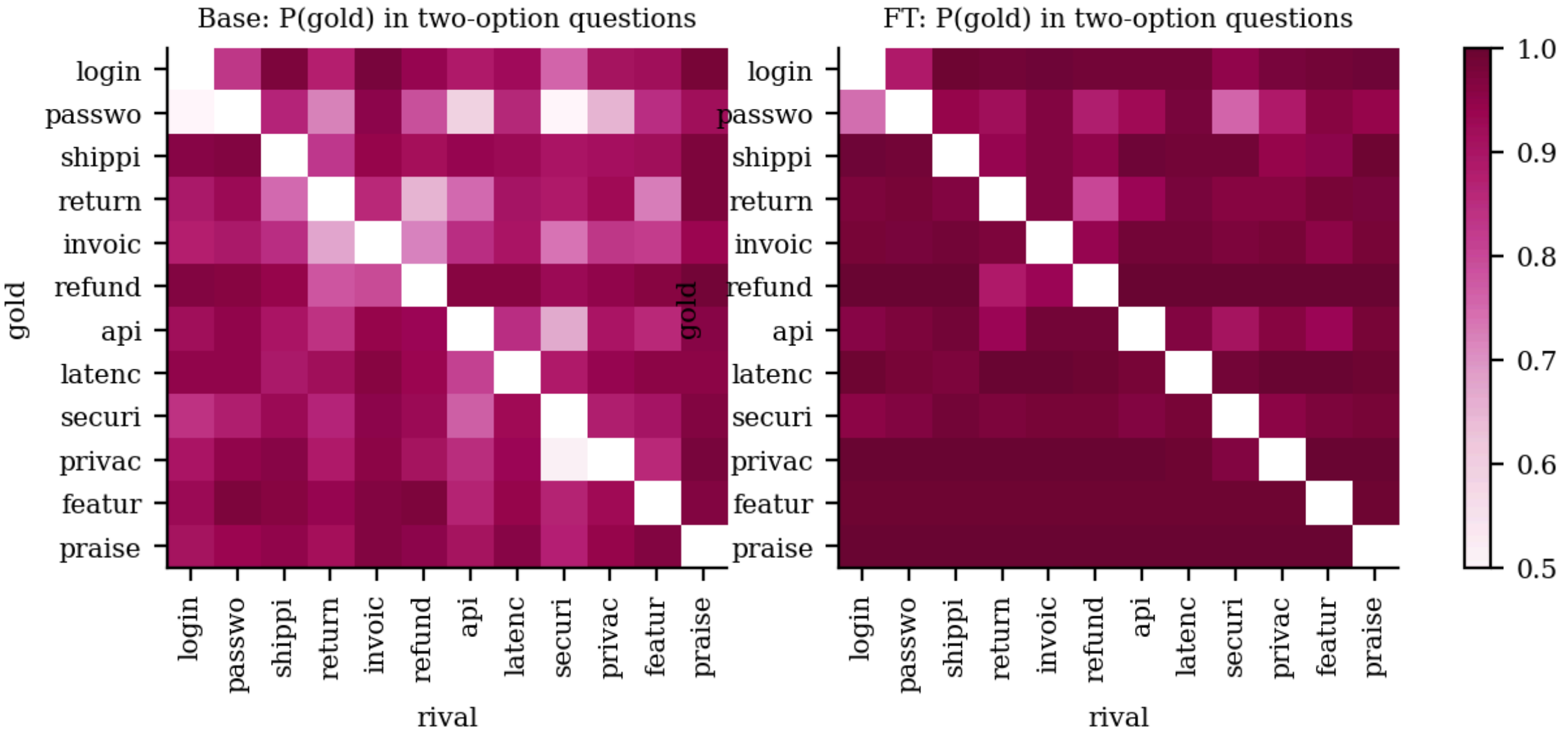

**Figure 9.** Pairwise confusion on the topic schema: mean probability that the gold label (row) beats the rival (column) in a two-option question. Pale cells are the confusable pairs.

## 7 The slot of the gold option

The gold option sits in a uniformly random slot of every question, so any dependence of accuracy on the slot belongs to the model. The topic and HR questions give 201 to 255 cases per slot at K=12 and Pool-24 gives 81 to 122 per slot at K=24 (Table 10, Figure 10). The pre-stated test, a spread of slot accuracies above a permutation null, passes in 1 of 6 cells (Pool-24 FT K=24; its spread of 21.4 points is only 0.7 points above the null) and fails in the others. P4 is therefore refuted as stated: with 81 to 255 cases per slot, the slot-to-slot differences are mostly of the size that chance produces.

A trend test has more power against a smooth primacy effect than a spread, and we ran it afterwards, so it is exploratory. It is nominally significant ($p < 0.05$) in 4 of 6 cells (HR-12 Base, HR-12 FT, Pool-24 Base, Pool-24 FT), always with earlier slots more accurate, and the fitted change from the first to the last slot is -8.1 to -6.4 points in those cells. With six tests and no correction, the safest reading is a primacy effect of several points on HR-12 and Pool-24, which FT did not or only partly saw, and none on the topic schema. The wrong answers do not cluster in one slot (Figure 11). The primacy effect is a sizeable part of the loss from K itself (11.2 points for FT on HR-12 between K=2 and K=12, 13.8 for Base), so it is worth shuffling the options when a deployment can.

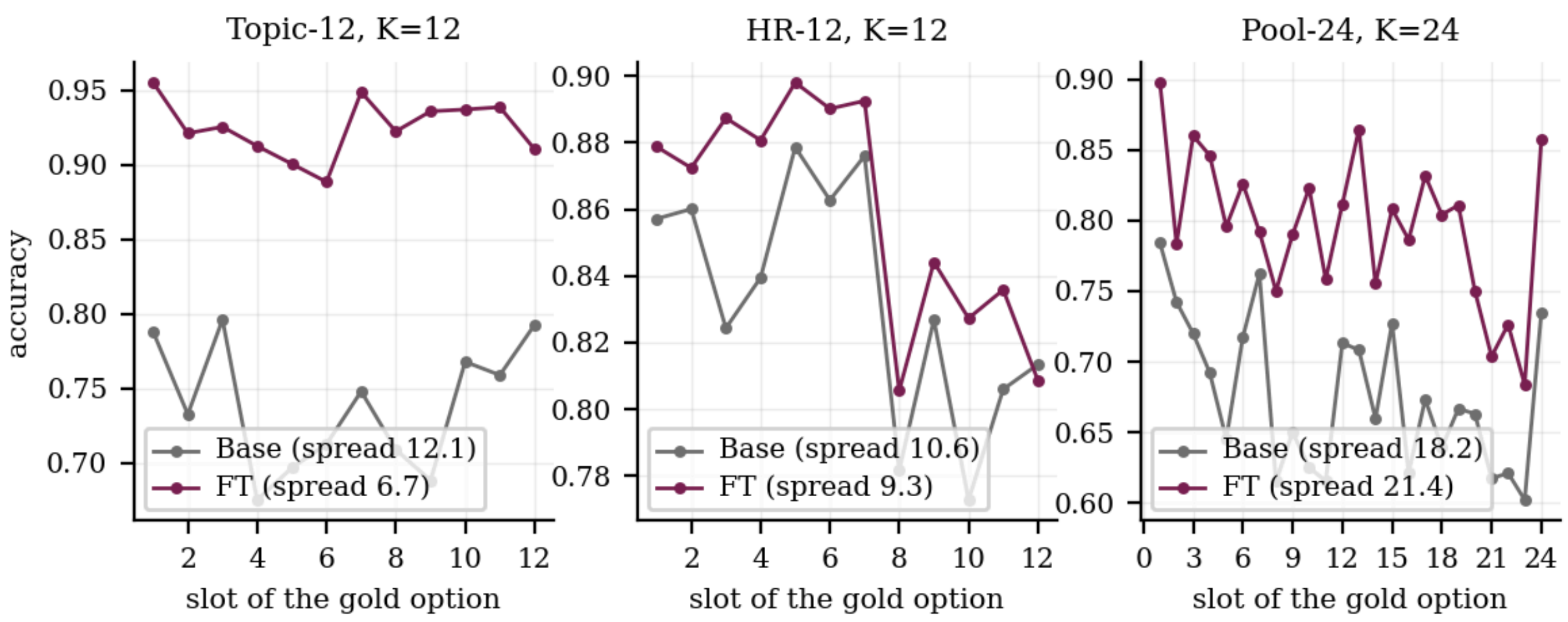

**Figure 10.** Accuracy by the slot in which the gold option is listed (slot 1 is first). The spread is the highest slot accuracy minus the lowest; each point pools the draws that put the gold option in that slot.

**Table 10.** Slot effect of the gold option. Spread: highest minus lowest slot accuracy (points). Null 95%: 95th percentile of the spread when the slot labels are shuffled (2,000 permutations); p: share of permutations with a spread at least as large. r: correlation between slot number and slot accuracy, with a permutation p for its absolute value (exploratory, not pre-stated). Change: fitted change in accuracy from the first to the last slot (points). n per slot: smallest and largest number of (item, draw) cases.

| Pool, K, model | Spread | Null 95% | p (spread) | r | p (trend) | Change, slot 1 to last | n per slot |
|---|---|---|---|---|---|---|---|
| Topic-12 K=12 Base | 12.1 | 13.7 | 0.140 | +0.07 | 0.828 | +0.9 | 201 to 254 |
| Topic-12 K=12 FT | 6.7 | 8.1 | 0.227 | -0.01 | 0.985 | -0.0 | 201 to 254 |
| HR-12 K=12 Base | 10.6 | 11.5 | 0.106 | -0.60 | 0.043 | -6.4 | 201 to 255 |
| HR-12 K=12 FT | 9.3 | 10.7 | 0.172 | -0.74 | 0.009 | -7.6 | 201 to 255 |
| Pool-24 K=24 Base | 18.2 | 24.5 | 0.523 | -0.47 | 0.025 | -8.0 | 81 to 122 |
| Pool-24 K=24 FT | 21.4 | 20.7 | 0.033 | -0.48 | 0.019 | -8.1 | 81 to 122 |

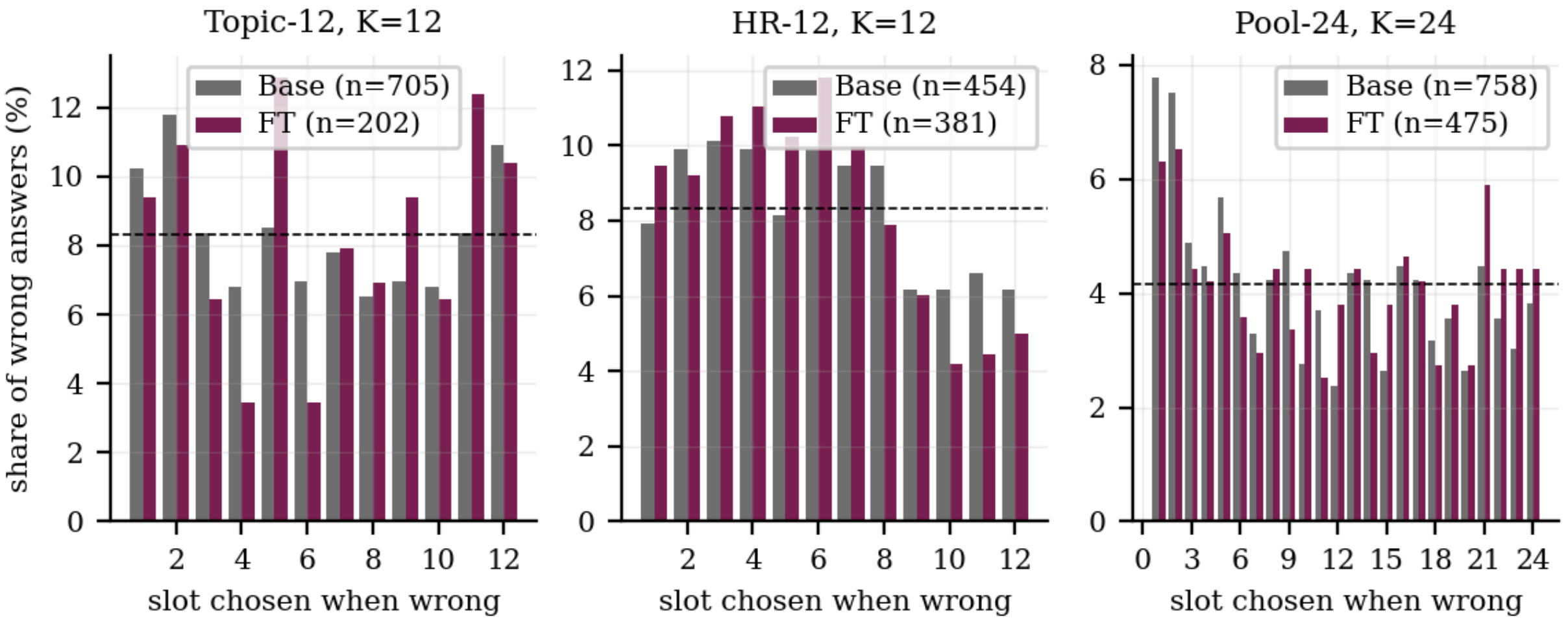

**Figure 11.** Where wrong answers land. Share of wrong answers that sit in each slot; the dashed line is the uniform share 1/K.

## 8 Fixes tested on held-out schemas

### 8.1 Averaging over option orders

Zhang et al. [2026] and Huang et al. [2026] average over cyclic rotations of the option list. Our nested draws give up to 10 orderings per item, so we can measure what averaging buys as the number of orderings grows (Figure 12, Table 11). On the topic schema FT gains +0.7 points and Base +1.9; on HR-12 the gains are +1.8 (FT) and +2.2 (Base); on Pool-24 at K=24, with 5 orderings, +1.3 (FT) and -0.4 (Base). No paired interval excludes zero, so the evidence for any gain is weak, and P6 (at least 2 points for Base) is refuted: Base reaches 2 points on HR-12 only. Dividing out the label prior, the other fix of Zhang et al., changes accuracy by -1.5 to +0.6 points, with every interval including zero. The ten-order average costs ten encoder passes per question, which at K=12 is ten times the cost of the original decision.

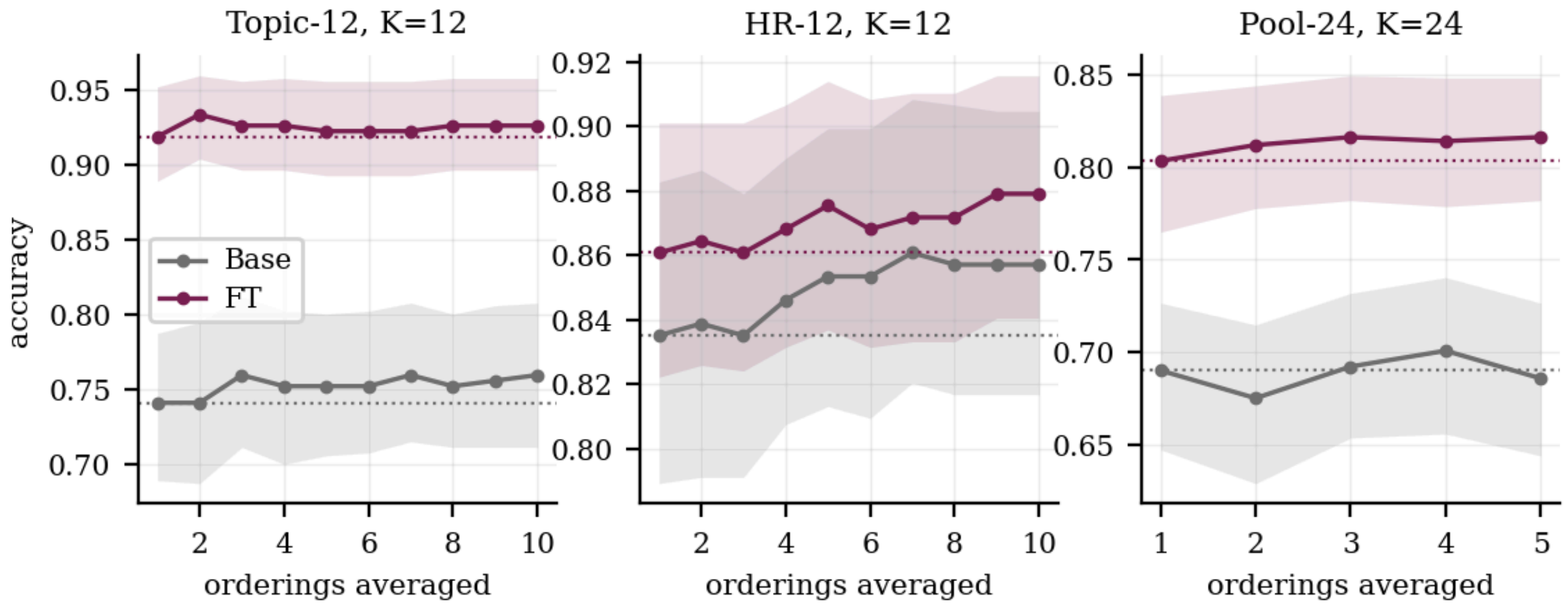

**Figure 12.** Permutation averaging. Accuracy of the mean probability over the first m random orderings of the full option list (solid, with 95% bands); the dotted line is one ordering.

**Table 11.** Two training-free fixes at the full option list. Orderings: mean probability over all stored orderings against one ordering. Prior division: the probability divided by the label's mean probability over the unlabelled test items. Both gains are paired differences to the single-ordering accuracy with 95% intervals over items. Flip: share of items whose answer changes between two random orderings.

| Pool, model | K | One ordering | Averaged | Gain, orderings | Gain, prior division | Flip (%) |
|---|---|---|---|---|---|---|
| Topic-12 Base | 12 | 74.1 | 75.9 (10 orders) | +1.9 [-1.1 to 5.2] | +0.4 [-1.1 to 2.2] | 12 |
| Topic-12 FT | 12 | 91.9 | 92.6 (10 orders) | +0.7 [-0.7 to 2.2] | -0.4 [-1.5 to 0.7] | 2 |
| HR-12 Base | 12 | 83.5 | 85.7 (10 orders) | +2.2 [0.0 to 4.4] | -0.4 [-1.5 to 0.7] | 8 |
| HR-12 FT | 12 | 86.1 | 87.9 (10 orders) | +1.8 [-0.4 to 4.0] | -0.4 [-2.2 to 1.1] | 8 |
| Pool-24 Base | 24 | 69.0 | 68.6 (5 orders) | -0.4 [-3.0 to 1.9] | -1.5 [-3.0 to 0.0] | 19 |
| Pool-24 FT | 24 | 80.3 | 81.6 (5 orders) | +1.3 [-0.9 to 3.4] | +0.6 [-0.6 to 2.1] | 11 |

### 8.2 Shortlist then compare

A listwise pass followed by a second pass on a shortlist is the structure of Runoff [Huang et al. 2026], which builds the shortlist by scoring each option alone. We build it from the listwise scores, so this tests the second stage only (Table 22, Table 12). The best shortlist size gains +0.0 to +2.6 points over a single pass across the twelve cells, and every interval includes zero. The best m is picked on the test items, so even these gains are optimistic. P7 predicted at least 3 points for Base and is refuted. The second stage asks a smaller question that the encoder answers well alone (two-option accuracy is 96 to 99% in the 12- and 24-label pools), but the smallest shortlist (two options in the 12-label pools, three in Pool-24) contains the gold option in only 90 to 99% of the items, so the second stage has little headroom.

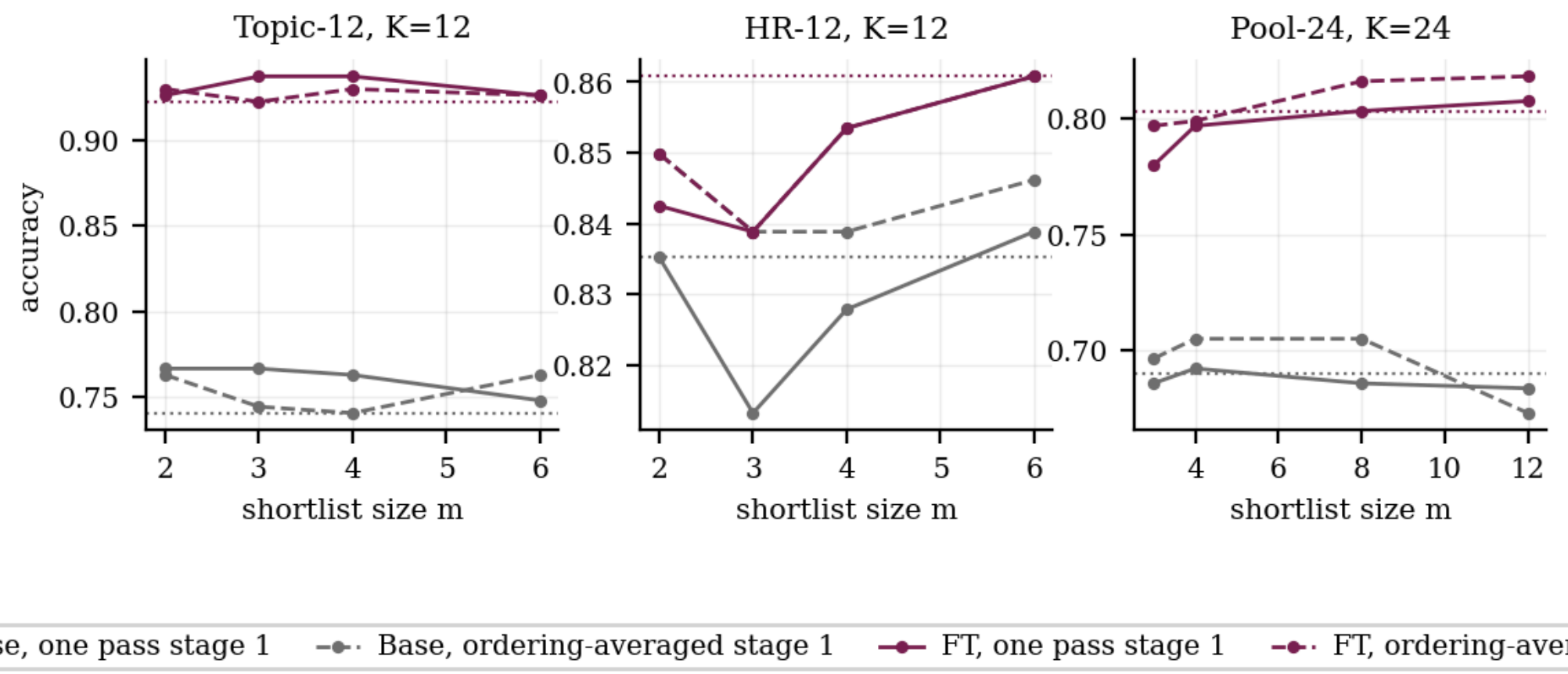

**Figure 13.** Shortlist then compare. The encoder first answers the full question; the top m options are then put to the same encoder as an m-option question. Solid: shortlist from a single pass; dashed: shortlist from the ordering-averaged probabilities; dotted: single-pass accuracy without the second stage.

**Table 12.** Shortlist then compare: the best shortlist size m for each pool, model and stage-1 source, with its paired gain over a single pass. Best m is chosen on the test items themselves, so the gain is optimistic by the selection (four sizes). The last column is the smallest shortlist.

| Pool, model | Stage 1 | Single pass | Best m | Gain [95% CI] | Smallest m |
|---|---|---|---|---|---|
| Topic-12 Base | one pass | 74.1 | m=2: 76.7 | +2.6 [-1.5 to 7.0] | m=2: 76.7 |
| Topic-12 Base | orderings | 74.1 | m=2: 76.3 | +2.2 [-1.9 to 7.6] | m=2: 76.3 |
| Topic-12 FT | one pass | 92.2 | m=3: 93.7 | +1.5 [-0.7 to 3.7] | m=2: 92.6 |
| Topic-12 FT | orderings | 92.2 | m=2: 93.0 | +0.7 [-2.0 to 3.0] | m=2: 93.0 |
| HR-12 Base | one pass | 83.5 | m=6: 83.9 | +0.4 [-3.7 to 3.7] | m=2: 83.5 |
| HR-12 Base | orderings | 83.5 | m=2: 85.0 | +1.5 [-2.8 to 6.2] | m=2: 85.0 |
| HR-12 FT | one pass | 86.1 | m=6: 86.1 | +0.0 [-3.3 to 2.9] | m=2: 84.2 |
| HR-12 FT | orderings | 86.1 | m=6: 86.1 | +0.0 [-3.5 to 2.9] | m=2: 85.0 |
| Pool-24 Base | one pass | 69.0 | m=4: 69.2 | +0.2 [-3.2 to 3.4] | m=3: 68.6 |
| Pool-24 Base | orderings | 69.0 | m=4: 70.5 | +1.5 [-2.4 to 5.1] | m=3: 69.7 |
| Pool-24 FT | one pass | 80.3 | m=12: 80.8 | +0.4 [-2.2 to 3.0] | m=3: 78.0 |
| Pool-24 FT | orderings | 80.3 | m=12: 81.8 | +1.5 [-1.8 to 4.3] | m=3: 79.7 |

## 8.3 A temperature law that transfers

The first part of this report found that the shipped temperature for 11 or more options is clamped to 0.5 (Section 15). Here we ask a different question. If one fits an extra multiplier as a function of K on one schema, does it help on another? Figure 14 and Table 13 answer for Topic-12 (the fit), HR-12 (a different schema) and Pool-24 (a larger K). On HR-12 at K from 8 to 12 the law changes the ECE by Base -9.2, FT +0.2 points (positive is better). It makes Base worse on HR-12 and leaves FT unchanged, because the best multiplier for FT is close to 1 for every pool and K, while the best multiplier for Base on Topic-12 at K=12 is 4.0 against 1.6 on HR-12. The law helps on Pool-24 at K=24 for both models. P8 is refuted: a temperature fitted on one schema does not transfer reliably to another, and the Base probabilities need a schema-specific fit.

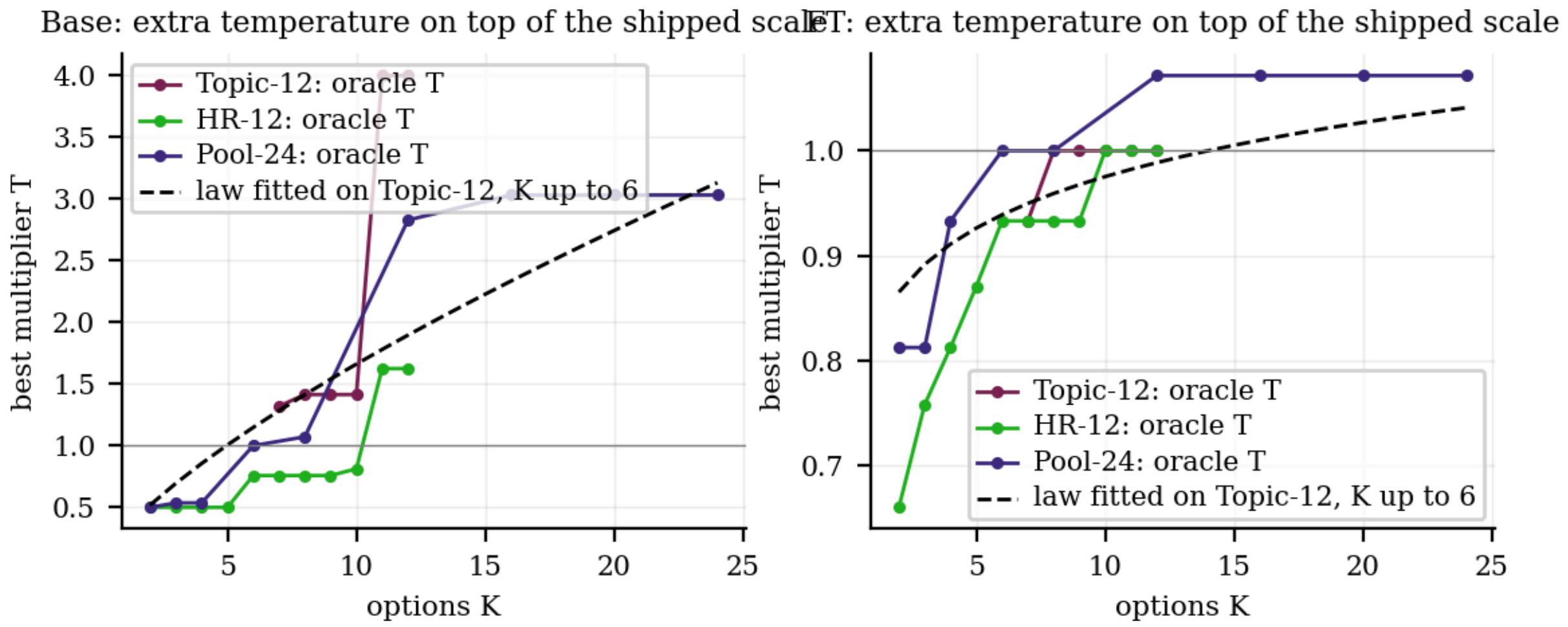

**Figure 14.** Is the temperature a function of K that transfers? Colored: the multiplier that minimises the negative log-likelihood on each pool and K (an oracle that sees the test labels). Dashed: log T linear in log K, fitted on Topic-12 at K up to 6 only. 1.0 means the shipped probabilities are already best.

**Table 13.** ECE (points, 15 bins) of the shipped probabilities, after the temperature law fitted on Topic-12 and after the oracle multiplier for that cell.

| Cell | ECE shipped | ECE law | ECE oracle | T law | T oracle |
|---|---|---|---|---|---|
| HR-12 K=6 Base | 6.7 | 10.0 | 3.7 | 1.15 | 0.76 |
| HR-12 K=12 Base | 8.4 | 9.2 | 7.1 | 1.90 | 1.62 |
| Pool-24 K=12 Base | 13.8 | 7.7 | 10.1 | 1.90 | 2.83 |
| Pool-24 K=24 Base | 22.4 | 15.0 | 14.7 | 3.13 | 3.03 |
| Topic-12 K=12 Base | 20.9 | 15.4 | 6.4 | 1.90 | 4.00 |
| HR-12 K=6 FT | 4.4 | 4.4 | 4.5 | 0.94 | 0.93 |
| HR-12 K=12 FT | 5.4 | 5.5 | 5.4 | 0.99 | 1.00 |
| Pool-24 K=12 FT | 4.2 | 4.4 | 3.6 | 0.99 | 1.07 |
| Pool-24 K=24 FT | 7.2 | 6.1 | 5.6 | 1.04 | 1.07 |
| Topic-12 K=12 FT | 2.1 | 2.3 | 2.1 | 0.99 | 1.00 |

## 8.4 Conformal sets across a schema shift

Kumar et al. [2023] study conformal prediction for multiple-choice questions and examine out-of-subject questions, where the exchangeability that the guarantee needs can fail. We calibrate a 90% set on half of the topic items and test it on the other half and on HR-12 (Figure 15). At K=12 the coverage on the held-out topic items is Base 96.5%, FT 94.6%, and on HR-12 Base 98.6%, FT 84.1%. FT under-covers on HR-12 by 6 points at K=12 and by more at smaller K. Base covers too much, with a mean set of 5.9 options out of 12, which is of little use. Recalibrating on 60 HR items and testing on the other 213 brings coverage to Base 92.1%, FT 87.5%. P9 asked for coverage within 2 points of the target on HR without recalibration and is refuted.

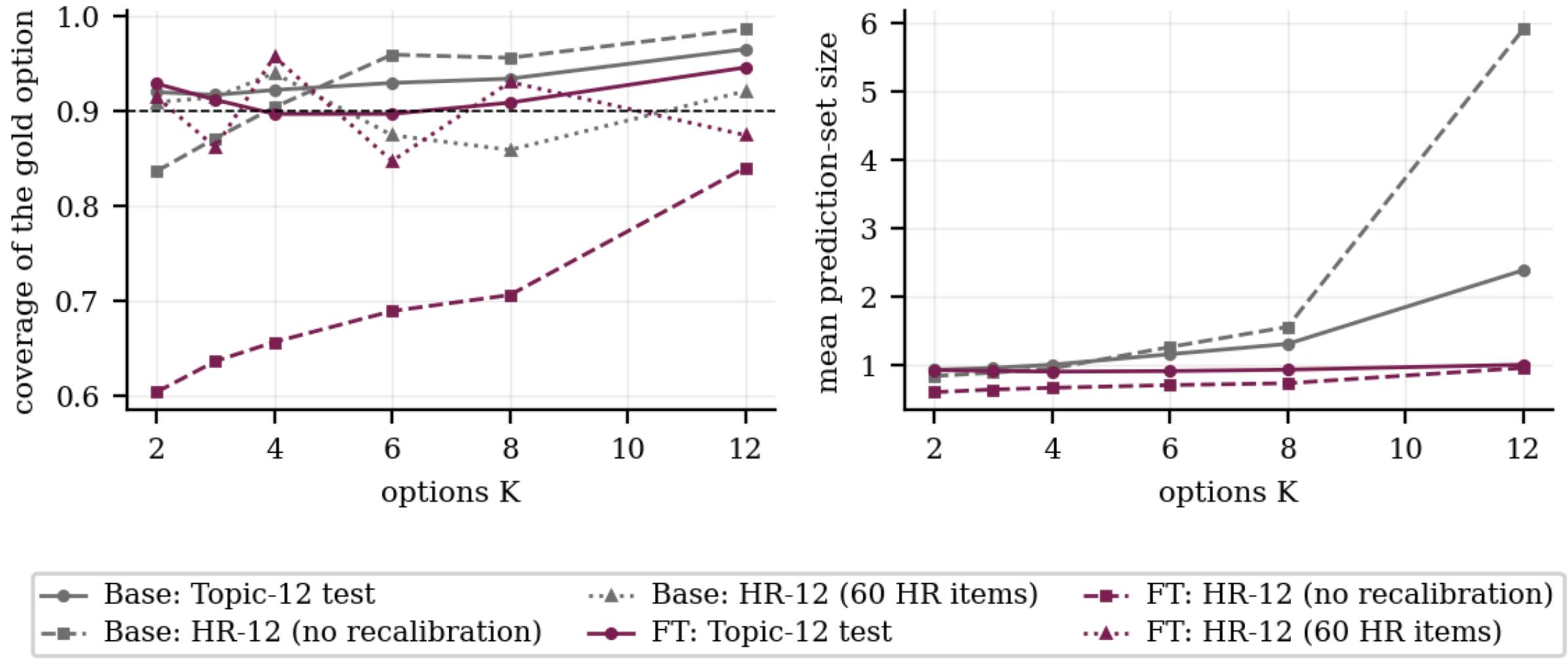

**Figure 15.** Split-conformal sets at a 90% target (threshold from half of the Topic-12 items, one draw per item). Left: coverage of the gold option on held-out topic items and on HR-12; the dotted line recalibrates on 60 HR items. Right: mean set size on both.

## 9 Thirty-five options and the head budget

Mega-35 goes to 35 options, far below the 255 that Rajput et al. [2026] reach on intent benchmarks. Accuracy at K=35 is Base 42%, FT 54%. The law fitted on K up to 6 of this pool would predict Base 50%, FT 69%, so it is too optimistic here (Figure 16). Families differ far more than in-family status explains (Table 14). For FT the topic items (in-family) keep 80% at K=35, but the HW and JSON items, also in-family, fall to 31% and 40%, below the HR items (56%) and level with the sentiment items (30%) that FT never saw. For Base the HW items reach 9% and the topic items 55%. A 35-option head with a description for each option is long, so the head budget could be cutting it. Raising it to 640 tokens (with a 1,024-token sequence) changes accuracy at K=35 by Base +0.8 [-0.8 to 2.5], FT +2.5 [0.1 to 4.4] points, so the cut explains little of the shortfall; P10, which expected a return to within 3 points of the K=24 curve, is refuted. We did not test whether the confusability of the 35 labels, which several domains share, explains the HW and JSON collapse.

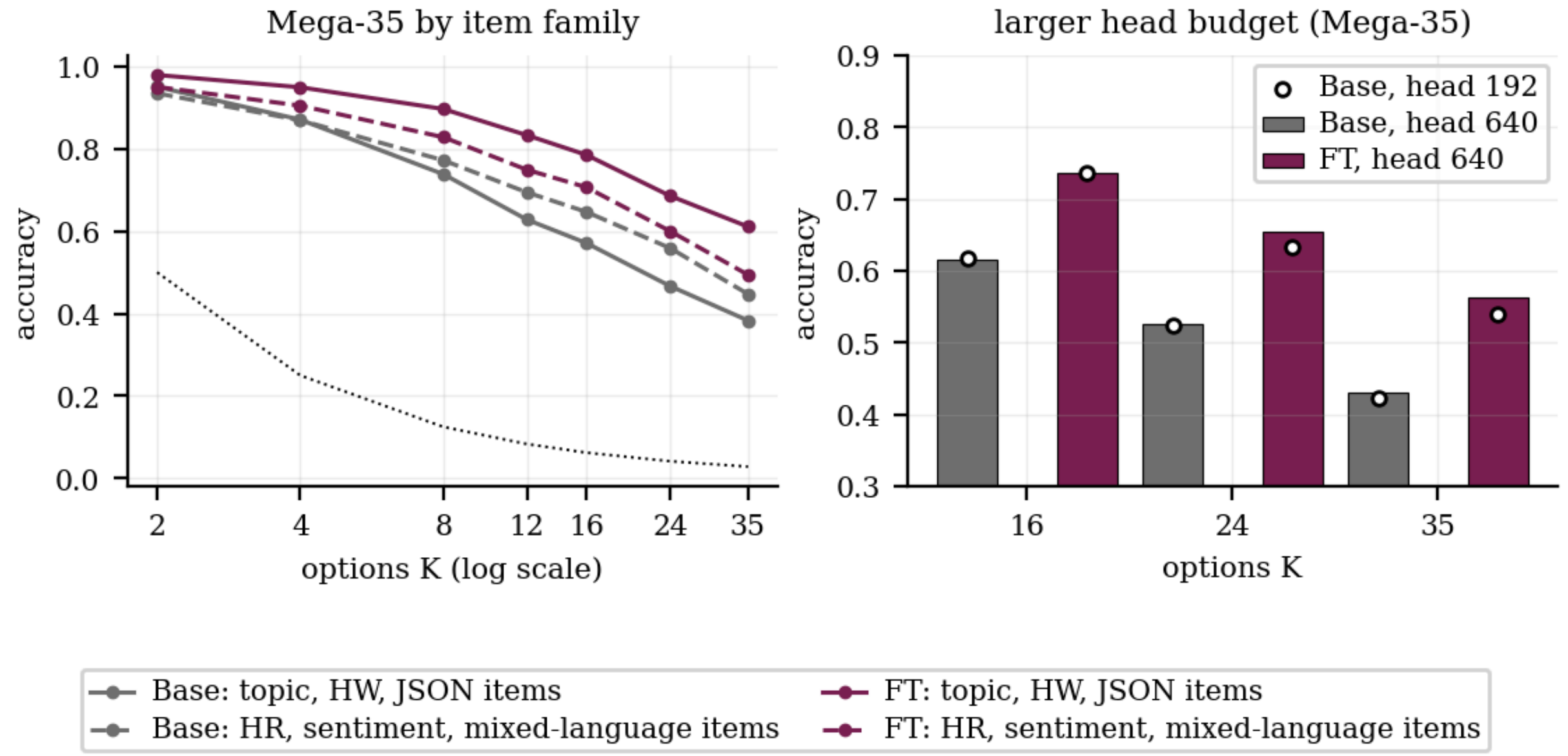

**Figure 16.** Left: accuracy on the 35-label pool by family of item. Right: the same items with the default head budget (open dots) and a head budget of 640 tokens with a 1,024-token sequence (bars).

**Table 14.** Accuracy (%) of the 35-label pool by source of the item. In-family means that the schema was in the fine-tuning data, which applies to FT only.

| Model | Source | FT in-family | n | K=2 | K=12 | K=24 | K=35 |
|---|---|---|---|---|---|---|---|
| Base | hr | no | 273 | 95 | 77 | 66 | 54 |
| Base | hw | yes | 82 | 89 | 45 | 24 | 9 |
| Base | json | yes | 58 | 95 | 45 | 28 | 25 |
| Base | many | yes | 195 | 97 | 76 | 62 | 55 |
| Base | mix | no | 156 | 90 | 66 | 50 | 41 |
| Base | sent | no | 106 | 95 | 54 | 37 | 27 |
| FT | hr | no | 273 | 95 | 80 | 67 | 56 |
| FT | hw | yes | 82 | 98 | 84 | 58 | 31 |
| FT | json | yes | 58 | 97 | 59 | 39 | 40 |
| FT | many | yes | 195 | 98 | 90 | 82 | 80 |
| FT | mix | no | 156 | 96 | 78 | 61 | 51 |
| FT | sent | no | 106 | 94 | 58 | 40 | 30 |

## 10 Scorecard of the predictions

Table 15 lists every pre-stated prediction. 2 of 10 held (P3, P5; P3 only up to K=24), 1 held in its first half (P2: IIA is rejected, but the sign and the model ordering we predicted were wrong), and the other 7 (P1, P4, P6, P7, P8, P9, P10) were refuted. The refuted ones include the predictions about position (P4) and order averaging (P6), the three about fixes (P7 shortlists, P8 a temperature law, P9 conformal sets), the composition test (P1) and the head budget (P10). The predictions that held concern structure, namely how accuracy depends on K and how the strongest rival causes the errors. Those regularities hold across the schemas we tried up to K=24, while procedures that adjust probabilities or answers did not carry over to a new schema.

**Table 15.** Scorecard of the ten pre-stated predictions. The middle column gives what the runs measured.

| ID | Prediction | Measured | Verdict |
|---|---|---|---|
| P1 | IIA composition within 3 points for FT, beyond 3 for Base | FT -0.7, -3.3, -4.7; Base +0.9, -2.8, -2.2 (largest K; topic, HR, Pool-24) | **refuted** |
| P2 | slope of the gold-vs-rival log-odds is negative for Base and smaller for FT; IIA rejected | slopes +0.24 to +2.07 nats per doubling; positive in every cell, larger for FT | IIA rejected; predicted sign and size refuted |
| P3 | held-out error under 2 points for K up to 12; breaks far beyond the fit | largest near error 0.9 points; Pool-24 K=24 0.7 (FT) and 1.6 (Base); Mega-35 11.1 (FT) and 5.4 (Base) | **confirmed to K=24; fails at K=35** |
| P4 | spread over slots above the permutation null | 1 of 6 cells with $p < 0.05$ | **refuted** |
| P5 | strongest rival predicts errors better than the count of rivals | AUROC 0.83 to 0.96 against 0.75 to 0.84; higher in 8 of 8 cells | **confirmed** |
| P6 | 10-order averaging gives Base at least 2 points | Base gains +1.9, +2.2, -0.4 (topic, HR, Pool-24); interval lower bounds -1.1, +0.0, -3.0 | **refuted** |
| P7 | shortlist-then-compare gives Base at least 3 points at the full K | best-m gain +2.6, +0.4, +0.2 (topic, HR, Pool-24; selected on the test items) | **refuted** |
| P8 | temperature law fitted on Topic-12 lowers ECE on HR | ECE change at K 8 to 12 on HR: Base -9.2, FT +0.2 points (positive is better) | **refuted** |
| P9 | conformal coverage within 2 points of 90% on HR | coverage at K=12 on HR Base 98.6%, FT 84.1% | **refuted** |
| P10 | head budget 640 restores K=35 to within 3 points of the K=24 curve | K=35 at head 640 against K=24 at head 192: Base 43.0% against 52.3%, FT 56.4% against 63.3% | **refuted** |

## 11 Threshold tests: scope and outcome

This part documents twenty test cases (T-01…T-20) run against Laya, a non-autoregressive encoder that answers a multiple-choice question about a text in one forward pass, with the number of options K pushed from 2 up to 24. They are the tests of the first version of this report, kept because they are what a practitioner can run in an afternoon; the measured laws of Sections 5 to 9 replace the thresholds as the main result. The tests cover accuracy as K grows, the confidence the runtime reports, the token budget that decides how much of each option the encoder reads, and sensitivity to the order in which options are listed. They also ask whether a data fix made for one 12-label schema carries over to another. Every test has a stated purpose, setup, acceptance threshold and rationale for that threshold; the observed value, its interval and the verdict are computed from frozen runs when this document is built.

Earlier work on typed-decision models has already reported order and position effects [Zhang et al. 2026; Teke 2026], label dominance over definitions [Azizi et al. 2026] and the use of temperature scaling [Guo et al. 2017]. Section 2 sets out what those papers show. This part adds a twenty-test matrix for Laya from K=2 to K=24, with a rationale for every threshold and the range over which each verdict holds. The head budget is accounted for token by token, which shows that the collapsed-option flag stays silent when only description text is lost (T-05 to T-10). The option-count temperature is stored as 0.1006 and applied as 0.5 (T-03, T-04). A fine-tune is tested with a second schema, four seeds and a TF-IDF baseline (T-13 to T-18).

Of the 14 tests that apply to the released checkpoint, 5 pass, 8 fail and 1 inconclusive; for the fine-tuned model (20 applicable tests) the counts are 13 pass, 4 fail and 3 inconclusive. The released model keeps 73.7% accuracy at K=12 against 95.2% at K=2 while reporting 95.2% confidence, and its stored 11+ temperature (0.1006) is applied as 0.5. Short descriptions at K=24 fit the default head budget (185 of 192 tokens). Long descriptions behind opaque option names do not: accuracy sits at 2.6-4.9% (chance 4.2%) and the runtime's collapsed-option flag stays silent in 100% of those questions.

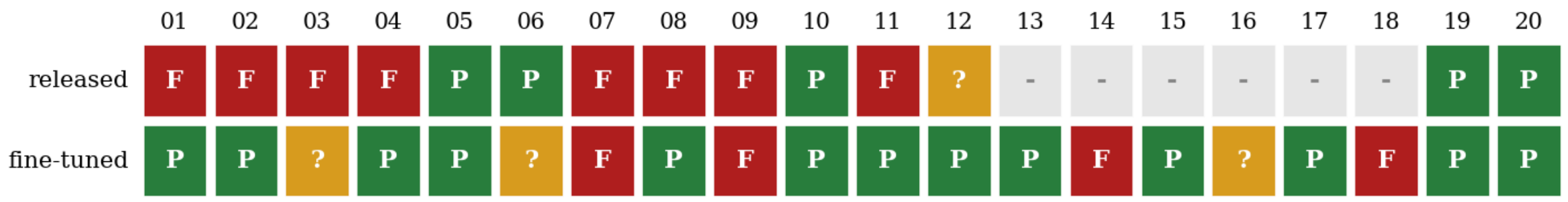

**Figure 17.** Verdict strip of the twenty test cases (P pass, F fail, ? inconclusive, - not applicable). Top row: released root checkpoint (T-03, T-09, T-10 and T-19 describe the runtime or the harness rather than a model). Bottom row: our fine-tune. Test numbers on top; the matrix in Section 12 gives the criteria.

The fine-tuned model passes most of the scaling and ordering tests. Its gain on a second 12-label schema is small: +1.5 points over the better released checkpoint (p=0.541), against +6.6 points (p=0.002) when the rows of that schema are added. All labels are written by an LLM, all text is synthetic, and a TF-IDF + logistic-regression model trained on the same rows is within 0.7 points of the best Laya variant on the HR schema. Section 22 lists what these results cannot show.

## 12 Test matrix

Observed values are given as *released root checkpoint* / *fine-tuned model* where both apply. Tests whose subject is the fine-tune itself (T-13…T-18) have a single observed value. The rationale under each threshold says why that number was chosen, and each test name states its threshold so that a verdict claims no more than the threshold measures.

**Table 16.** Test matrix. Verdicts are computed from the observed value and its 95% interval against the stated threshold.

| ID | Test, threshold and rationale | Observed | Released | Fine-tuned |
|---|---|---|---|---|

| ID | Test, threshold and rationale | Observed | Released | Fine-tuned |
|---|---|---|---|---|
| T-01 | **At K=12, accuracy keeps 90% of its K=2 value**<br>12-topic question; K=2 vs K=12 options sampled around the gold topic; 270 items × 3 draws; cluster bootstrap over items<br>Threshold: acc(K=12) / acc(K=2) ≥ 0.90, lower 95% bound.<br>*Why: 0.90 is the largest relative loss we would accept when a two-option form becomes a twelve-option one. It is a round number, and the lower bound must clear it.* | 0.77 [0.72-0.82] / 0.94 [0.91-0.96] | FAIL | PASS |
| T-02 | **Over-confidence at K=12 stays within 5 points**<br>same draws; mean top probability minus accuracy<br>Threshold: gap ≤ +5 points, upper 95% bound.<br>*Why: 5 points is the effect size used for the accuracy tests. Only over-confidence is tested: it breaks a confidence gate, while under-confidence only wastes answers.* | +21.5 [17.3-26.1] / +1.7 [-1.1 to 4.4] | FAIL | PASS |
| T-03 | **Stored temperature is the one applied**<br>static: read `temperature_by_options[choice:11+]` and pass it through the runtime's `clamp_temperature`<br>Threshold: applied = stored (±5%), and not pinned to a clamp bound.<br>*Why: 5% absorbs rounding in the stored value. A value on a clamp bound cannot be read as fitted, because the fit may have wanted to go further.* | 0.1006 → 0.50 / choice type temperature 5.0 → 5.0 (config has no per-bucket table) | FAIL | INCONCLUSIVE |
| T-04 | **Applied temperature is within 0.03 ECE of the optimum**<br>270 twelve-option items; re-scale the runtime's probabilities by a multiplier m; compare ECE at m=1 with the best m on a grid<br>Threshold: ECE(m=1) – min ECE ≤ 0.03, upper 95% bound.<br>*Why: On 270 items a perfectly calibrated predictor already has ECE of about 0.024 (simulated, ten bins), so 0.03 keeps the criterion near sampling noise.* | 0.179 [0.098-0.230] / 0.000 [0.000-0.023] | FAIL | PASS |
| T-05 | **No item changes with the head budget (short descriptions)**<br>24 options with label-only / short descriptions; `head_max_len` 192 vs 768; 468 items<br>Threshold: accuracy unchanged (no item flips).<br>*Why: With short descriptions the head fits the default budget, so the token sequences are identical and any change would be a harness fault. No tolerance is needed.* | 0 of 468 items change / 0 of 468 | PASS | PASS |
| T-06 | **Log-linear trend predicts K=24 within 5 points**<br>per-doubling slope fitted on K=2..12 (12-topic question) added to the pooled K=12 accuracy of the 24-label pool; compare with measured K=24 (468 items)<br>Threshold: \|measured – predicted\| ≤ 5 points.<br>*Why: 5 points matches the effect criterion. One K=24 accuracy at 468 items already has a half-width of 4.2 points, so this band is near the resolution limit and can end inconclusive.* | pred 70.2, meas 69.2 (-1.0) / pred 83.8, meas 79.1 (-4.7) | PASS | INCONCLUSIVE |
| T-07 | **Long descriptions at the default budget beat twice chance**<br>24 opaque names (`x0..x23`), 28-34-token descriptions, informative words last; `head_max_len` 192; 468 items<br>Threshold: accuracy ≥ 2× chance (2×4.2% = 8.3%), lower 95% bound.<br>*Why: Twice chance is the lowest bar that separates reading the options from guessing. It is a floor and does not show that the output is usable.* | 4.9% [3.3-7.3] / 4.1% [2.6-6.3] | FAIL | FAIL |

| ID | Test, threshold and rationale | Observed | Released | Fine-tuned |
|---|---|---|---|---|
| T-08 | **Raised budget lifts long-description accuracy to 25%**<br>same question; `head_max_len` 768 with `max_len` raised to 1152<br>Threshold: accuracy ≥ 25%, lower 95% bound.<br>*Why: A quarter is a round number about six times chance (4.2%). It sits well above the T-07 floor and well below the 79% the fine-tune reaches with short options at K=24.* | 16.7% [13.6-20.3] / 38.2% [34.0-42.7] | FAIL | PASS |
| T-09 | **Collapsed-options flag fires on 95% of questions that lose their description text**<br>share of questions in T-07 for which `usage["options"]` reports collapsed options; the cut removes every informative token<br>Threshold: ≥ 95% flagged (lower 95% bound).<br>*Why: No label-specific token survives in this design, so a working guard should fire almost always. The remaining 5% absorbs measurement slack.* | 0.0% / 0.0% | FAIL | FAIL |
| T-10 | **Too small a `max_len` is refused, not truncated**<br>static: `build_sequence` with `head_max_len=768` and the default `max_len=512`; the runtime raises when the surviving marker count differs from the option count<br>Threshold: fewer markers than options (runtime would raise `ValueError`).<br>*Why: A deterministic read of the runtime's own functions, so no tolerance applies.* | 17 of 24 markers survive; guard condition met | PASS | PASS |
| T-11 | **Option order changes the outcome for at most 15% of items**<br>K=12; 24 random orderings per item (270 items); share of items that are right under some orderings and wrong under others<br>Threshold: flip share ≤ 15%, upper 95% bound.<br>*Why: Well under the 0.33 order-flip rate that Zhang et al. [2026] report for unmodified readouts on 20-option tasks, and above the 2.3-2.8% answer change of a shuffle-trained sequential head [Teke 2026]. The flip definitions differ, so the bar is lenient: a pass does not mean order independence.* | 31.5% [25.9-37.4] / 10.7% [7.4-14.4] | FAIL | PASS |
| T-12 | **No slot gets more than twice its uniform share**<br>K=12 options with identical descriptions, 120 unrelated states, random label order; largest mean slot probability relative to uniform (1/12)<br>Threshold: ≤ 2.0× uniform (inconclusive within 15% of the limit).<br>*Why: A slot with twice its uniform share dominates when the text is uninformative. The estimate has no interval (120 texts), so the verdict is inconclusive within 15% of the limit (1.7-2.3×).* | 2.17× / 1.54× | INCONCLUSIVE | PASS |
| T-13 | **In-schema: the fine-tune beats the released model by 5 points**<br>full 12-way topic question, 195 held-out items; exact McNemar<br>Threshold: ≥ +5 points and $p < 0.01$.<br>*Why: 5 points is about three times the seed sd of the fine-tune on topic items (1.6 points). p<0.01 is close to 0.05 divided by the five paired comparisons (T-13, T-14, T-15 and two in T-18).* | 73.3% → 92.8% (+19.5 pts, p<0.001) | n/a | PASS |
| T-14 | **Transfer: gain of at least 5 points on an unseen 12-label schema**<br>HR-request schema never used in training, 273 items; fine-tune vs best released (Base)<br>Threshold: ≥ +5 points and $p < 0.01$.<br>*Why: Same criterion as T-13.* | 82.8% → 84.2% (+1.5 pts, p=0.541) | n/a | FAIL |

| ID | Test, threshold and rationale | Observed | Released | Fine-tuned |
|---|---|---|---|---|
| T-15 | **Schema adaptation: HR rows add 5 points on HR**<br>FT vs FT+HR on the 273 HR items<br>Threshold: ≥ +5 points and $p < 0.01$.<br>*Why: Same criterion as T-13.* | 84.2% → 90.8% (+6.6 pts, p=0.002) | n/a | PASS |
| T-16 | **Topic accuracy stays within 2 points after adding HR rows**<br>FT vs FT+HR on the 195 topic items; paired bootstrap of the difference<br>Threshold: difference within ±2 points (95% interval inside).<br>*Why: 2 points is about the seed sd (1.6 on topic items, 0.7 on HR items); a change inside seed noise is not damage.* | +1.0 pts [-2.1 to 4.1] | n/a | INCONCLUSIVE |
| T-17 | **Seed sd of the fine-tune is at most 2 points**<br>K=12 accuracy of four trainer seeds (same data)<br>Threshold: sd ≤ 2 points (inconclusive up to 3).<br>*Why: At sd 2 points the 5-point effect bar is 2.5 sd; at the inconclusive limit of 3 points it is 1.7 sd.* | sd 1.6 pts; range 89.6-93.0% | n/a | PASS |
| T-18 | **Beats a bag-of-n-grams classifier by 5 points on the same rows**<br>TF-IDF + logistic regression on the training rows of each schema vs FT (topics) and FT+HR (HR); exact McNemar<br>Threshold: ≥ +5 points and $p < 0.01$ on both schemas.<br>*Why: Same criterion as T-13. The classifier is the cheap baseline a deployment would otherwise use.* | topics +4.6 pts (p=0.078); HR +0.7 pts (p=0.875) | n/a | FAIL |
| T-19 | **An independent LLM agrees with the labels on 90% of items**<br>independent LLM (gemma4-31b) labels every held-out item blind; agreement with the Qwen3.8-27B label<br>Threshold: agreement ≥ 90%, lower 95% bound, both schemas.<br>*Why: A convention. Up to 10% disagreement could move accuracy by up to 10 points, more than the 5-point criterion, so Section 19 repeats two of the effect tests (T-15 and T-16) on the agreed items.* | topics 94.4% [90.2-96.8]; HR 96.7% [93.9-98.3] | PASS | PASS |
| T-20 | **Negative control: no information gives chance**<br>K=12 with opaque names and empty descriptions (270 items); the answer cannot be inferred from the question<br>Threshold: chance (8.3%) inside the 95% interval.<br>*Why: With opaque names and no descriptions, an accuracy outside the interval around chance would mean the harness leaks the answer.* | 8.5% [5.7-12.5] / 9.3% [6.4-13.3] | PASS | PASS |

### 12.1 How far each threshold can move

Table 17 lists, for each test with an adjustable threshold, the range in which the verdict is unchanged. The verdicts closest to their limits are the released T-06 (a pass by 0.1 points of band width), the fine-tuned T-06 (4.7 points against a 5-point band) and the released T-12 (2.17× against 2.0×); read them as unsettled. The large effects (the released T-01, T-07 and T-09, and T-13) keep their verdicts unless the threshold moves past the observed value itself.

**Table 17.** Range of each threshold over which the verdict stays the same, from the observed value and its 95% bound (the interval rules of Section 3.3 run in reverse). T-03, T-05, T-10 and T-20 have no threshold to move. For effect tests, 'bar' is the minimum effect and α the p-value limit.

| ID | Threshold as set | Released | Fine-tuned |
|---|---|---|---|
| T-01 | ratio ≥ 0.90 | pass ≤ 0.72; fail > 0.77 | pass ≤ 0.91; fail > 0.94 |
| T-02 | gap ≤ 5 pts | pass ≥ 26.1 pts; fail < 21.5 pts | pass ≥ 4.4 pts; fail < 1.7 pts |
| T-04 | excess ≤ 0.03 | pass ≥ 0.230; fail < 0.179 | pass ≥ 0.023; fail < 0.000 |
| T-06 | ±5 pts | pass ≥ 4.9 pts; fail < 1.0 pts | pass ≥ 8.5 pts; fail < 4.7 pts |

| ID | Threshold as set | Released | Fine-tuned |
|---|---|---|---|
| T-07 | acc ≥ 8.3% | pass ≤ 3.3%; fail > 4.9% | pass ≤ 2.6%; fail > 4.1% |
| T-08 | acc ≥ 25% | pass ≤ 13.6%; fail > 16.7% | pass ≤ 34.0%; fail > 38.2% |
| T-09 | flagged ≥ 95% | pass ≤ 0.0%; fail > 0.0% | pass ≤ 0.0%; fail > 0.0% |
| T-11 | flips ≤ 15% | pass ≥ 37.4%; fail < 31.5% | pass ≥ 14.4%; fail < 10.7% |
| T-12 | ≤ 2.0× uniform | pass ≥ 2.55×; fail < 1.89× | pass ≥ 1.81×; fail < 1.34× |
| T-13 | ≥ 5 pts, p < 0.01 | n/a | pass if bar ≤ 19.5 pts and α > 0.001 |
| T-14 | ≥ 5 pts, p < 0.01 | n/a | pass if bar ≤ 1.5 pts and α > 0.541 |
| T-15 | ≥ 5 pts, p < 0.01 | n/a | pass if bar ≤ 6.6 pts and α > 0.002 |
| T-16 | ±2 pts | n/a | pass ≥ 4.1 pts; fail < 1.0 pts |
| T-17 | sd ≤ 2 pts | n/a | pass ≥ 1.6 pts; fail < 1.1 pts |
| T-18 | ≥ 5 pts, p < 0.01, both | n/a | pass if bar ≤ 0.7 pts and α > 0.875 |
| T-19 | agreement ≥ 90% | pass ≤ 90.2%; fail > 94.4% | same |

# 13 Token-budget accounting

## 13.1 The accounting rules

Three rules from `build_head` and `build_sequence` decide what the encoder sees. (1) Each option is first cut to 48 tokens. (2) If the options then leave fewer than 16 tokens of `head_max_len` for the instruction, every option is cut again to `max(4, (head_max_len-16)//K)` tokens and the instruction takes what remains (at least 8 tokens). (3) The text is cut from the right to the room that `max_len` leaves after the head. Rule 2 is the only place where K enters, and it acts on *every* option at once. A question with 24 options of 20 tokens already exceeds 192, so all 24 are shortened: the first words survive, and the last ones, where a long description usually states its distinguishing content, go first. Nothing is reported to the caller about rule 2 except through the collapsed-option flag, which fires only if two options become token-identical.

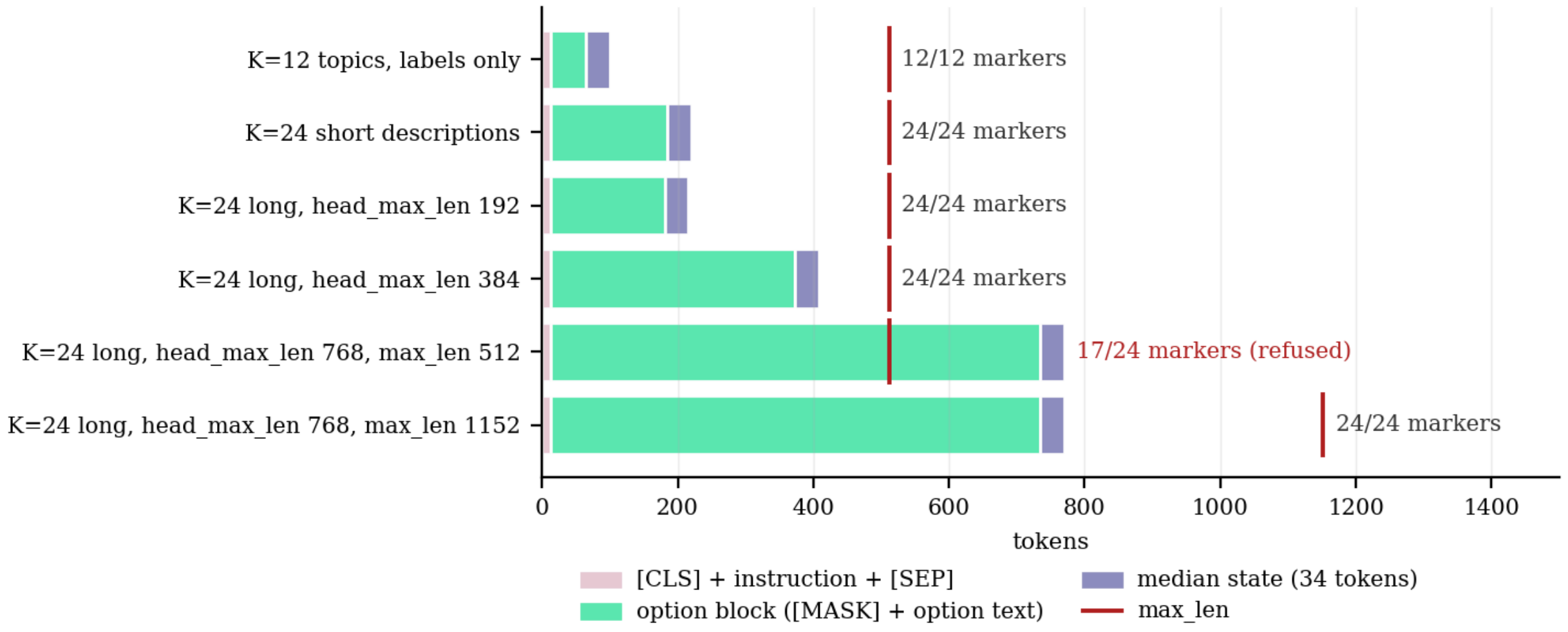

**Figure 18.** Token-budget stack drawn from the runtime's own `build_head` for six configurations. The red tick is `max_len`; markers past it are dropped and the runtime then refuses the request.

## 13.2 Where rule 2 starts to bind

The measured head of the 24 short options (label plus keyword description) is 185 tokens, i.e. 7 tokens below the default budget, so rule 2 does not fire and the 3-11-token options are read whole at any `head_max_len` ≥ 192. This is the static explanation of the null result in T-05. With longer descriptions the limit arrives early: at K=16 an 8-token description is already cut (cap 11), and at K=24 a 4-token description is cut to 7 tokens including the marker (see Figure 19, centre and right panels).

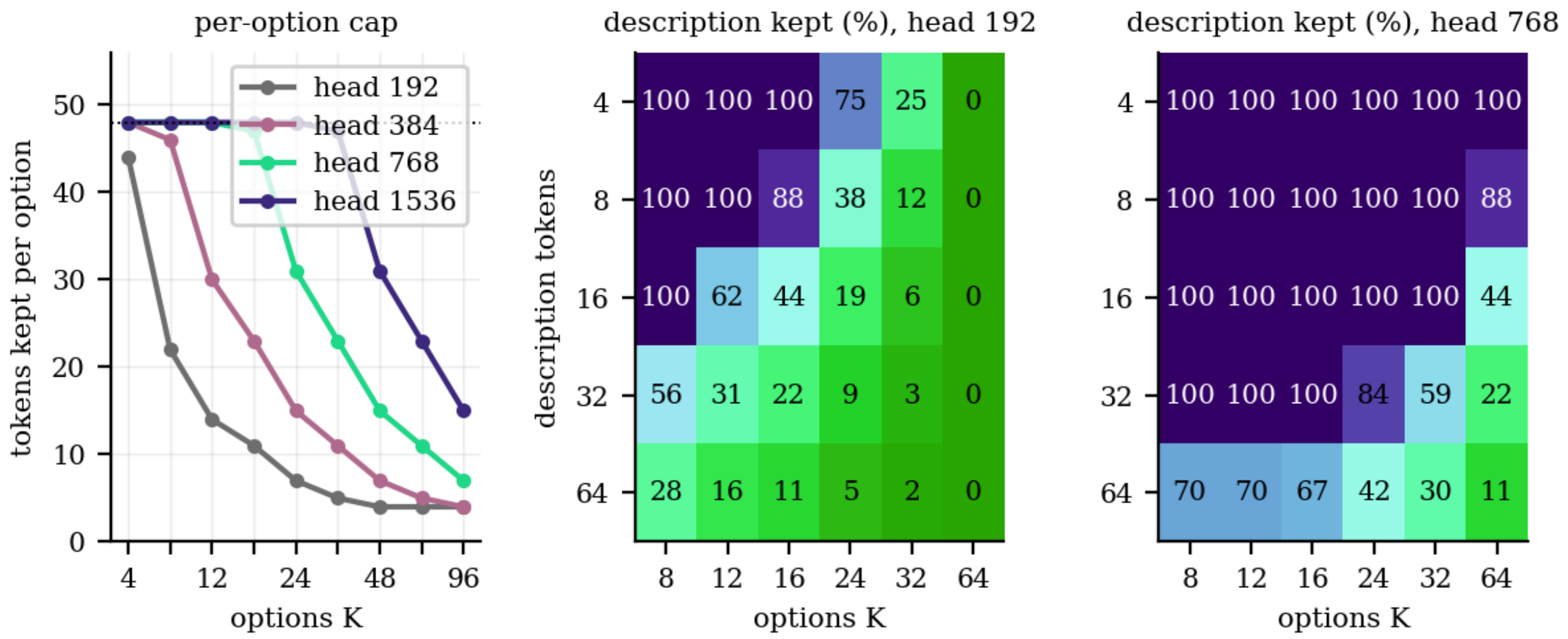

**Figure 19.** Left: tokens each option keeps as K grows when descriptions are long enough to hit the caps (60-token descriptions; hard cap 48, dotted). Centre and right: share (%) of an option's description tokens that survive the head budget, by number of options and description length, at the default budget and at 768.

### 13.3 Ledger for the stress case

The long-description question was designed so that cutting a description removes the information: each option is an opaque name (`x0…x23`) plus a 23-token preamble that is identical for all options plus the label-specific words (1-7 tokens). At the default budget the cap is 7 tokens, so the model sees a name, a colon and the first few words of the preamble; no informative token survives. At 384 the cap is 15: still inside the preamble. At 768 the cap is 31 and 57% of the informative tokens survive; at 1536 nothing is cut. Table 18 combines this static picture with the measured accuracy.

**Table 18.** Budget ledger for the 24-option long-description question (28-34 tokens per option, of which the first 27 (name, colon and a shared preamble) carry no label information and the last 1-7 do). 'Informative share' = label-specific tokens that survive. Right-hand columns: accuracy measured with `max_len` raised as shown (T-07, T-08).

| head_max_len | head tokens | cap/option | tokens kept (mean / full) | informative share | markers at max_len 512 | max_len used | Base | TD | FT | FT+HR |
|---|---|---|---|---|---|---|---|---|---|---|
| 192 | 181 | 7 | 6.0 / 30.6 | 0% | 24 / 24 | 576 | 5% | 3% | 4% | 4% |
| 384 | 373 | 15 | 14.0 / 30.6 | 0% | 24 / 24 | 768 | 4% | 4% | 4% | 4% |
| 768 | 735 | 31 | 29.1 / 30.6 | 57% | 17 / 24 | 1152 | 17% | 36% | 38% | 56% |
| 1536 | 772 | none | 30.6 / 30.6 | 100% | 17 / 24 | 1920 | 10% | 41% | 17% | 47% |

### 13.4 Which (head, sequence) pairs are legal

Raising `head_max_len` alone is not enough: with `max_len`=512 the 768-token head places 17 of 24 markers inside the sequence, so the runtime refuses the call (T-10, the one place where it fails loudly). The map also shows a quieter hazard: pairs such as (1024, 768) keep every marker but leave 0 tokens for the text, so the text is dropped entirely. The runtime does expose a truncation flag for the text in `usage`, but a caller who does not read it gets answers computed without the document. The frozen long-description runs avoided this region by setting `max_len=head_max_len+384`.

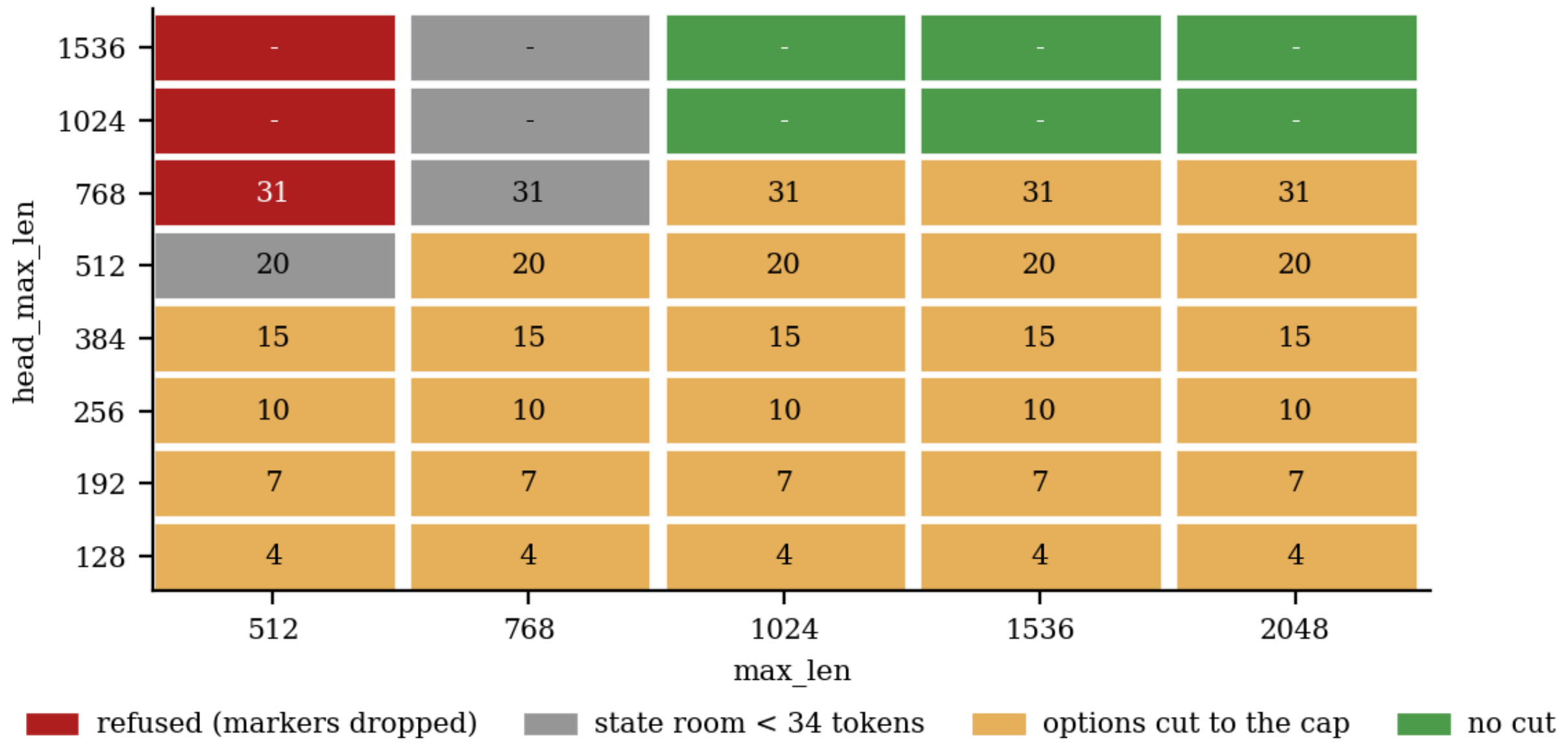

**Figure 20.** Feasibility map for the 24-option long-description question: for each (`head_max_len`, `max_len`) pair, whether the runtime would refuse, leave no room for the state, cut options, or fit. Numbers: tokens kept per option (= none cut).

## 14 Scaling curves in K

### 14.1 Accuracy

T-01 asks that a model keep 90% of its two-option accuracy at twelve options. The released root checkpoint keeps 0.77 [0.72-0.82] and fails; the fine-tune keeps 0.94 and passes. The loss is steady on a log scale, 8.2 points per doubling of K for Base against 2.6 for FT, and the multilingual sub-checkpoint loses 13.6. Because chance falls with K, the chance-adjusted curves on the right of Figure 21 are the fairer picture. Base's skill falls from 90% to 71%, FT's from 98% to 92%, so the decline is more than a shrinking chance floor.

**Table 19.** Accuracy (%) by number of options on the 12-topic question (270 items × 3 draws; full intervals in Appendix C) and on the HR schema (273 items × 3 draws; none of Base, TD, ML or FT was trained on it, FT+HR was). Slope: points per doubling of K.

| Schema: model | K=2 | K=3 | K=4 | K=6 | K=8 | K=12 | Slope / doubling | K=12 / K=2 |
|---|---|---|---|---|---|---|---|---|
| Topic: **Base** | 95.2 | 92.8 | 89.8 | 83.8 | 82.3 | 73.7 | -8.2 [-9.9 to -6.7] | 0.77 |
| Topic: **TD** | 94.2 | 92.6 | 89.8 | 84.2 | 80.4 | 74.2 | -8.0 [-9.8 to -6.3] | 0.79 |
| Topic: **ML** | 92.0 | 85.9 | 79.3 | 71.4 | 66.5 | 56.8 | -13.6 [-15.6 to -11.8] | 0.62 |
| Topic: **FT** | 99.0 | 98.8 | 98.9 | 95.7 | 95.4 | 92.6 | -2.6 [-3.6 to -1.6] | 0.94 |
| Topic: **FT+HR** | 99.0 | 98.3 | 97.7 | 95.7 | 94.9 | 93.0 | -2.4 [-3.5 to -1.4] | 0.94 |
| HR: **Base** | 97.6 | 94.3 | 93.0 | 90.4 | 86.1 | 83.4 | -5.5 | 0.85 |
| HR: **TD** | 97.3 | 95.0 | 92.6 | 90.7 | 85.0 | 82.1 | -6.1 | 0.84 |
| HR: **ML** | 86.4 | 80.8 | 75.6 | 70.9 | 63.5 | 54.9 | -12.0 | 0.64 |
| HR: **FT** | 97.9 | 94.9 | 93.5 | 90.8 | 88.3 | 85.5 | -4.8 | 0.87 |
| HR: **FT+HR** | 98.3 | 97.1 | 95.2 | 95.0 | 92.9 | 90.1 | -3.0 | 0.92 |

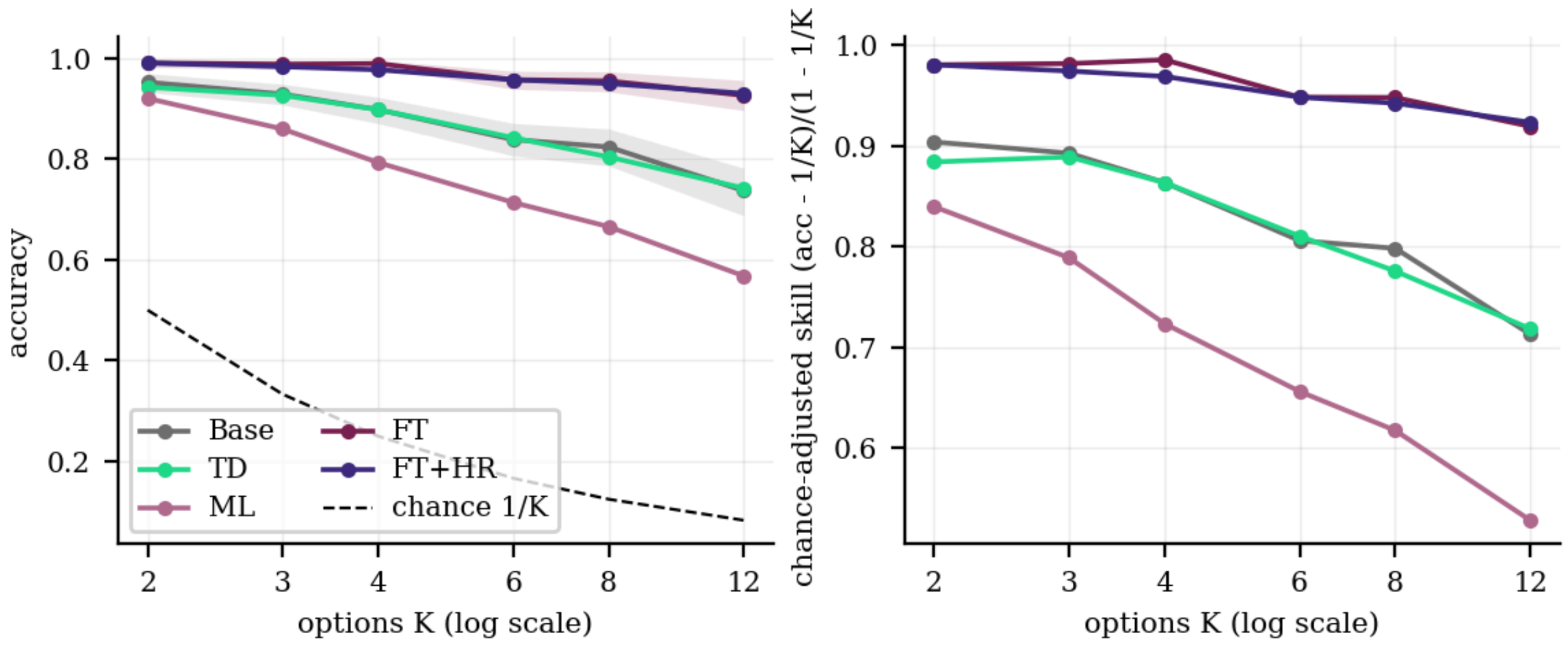

**Figure 21.** Scaling curves on the 12-topic question. Left: accuracy against K on a log2 axis (bands: 95% bootstrap over items for Base and FT). Right: the same curves after removing the chance floor; a flat line there would mean the model loses nothing but chance.

### 14.2 Confidence

For Base the top probability does not follow accuracy: it is 91% at K=2, dips to 84% at K=4 and climbs to 95% at K=12 while accuracy falls to 74% (Figure 22). At K=12 the over-confidence is +21.5 points [17.3-26.1] and T-02 fails; the fine-tune's gap is +1.7 [-1.1 to 4.4]. The shape follows the stored temperature buckets (Table 20): 1.91 for two options, 1.76 for three to five, 1.00 for six to ten and, above ten, the clamped 0.5 instead of the stored 0.1006. Confidence falls where the divisor softens and rises where it does not. This is the subject of Section 15.

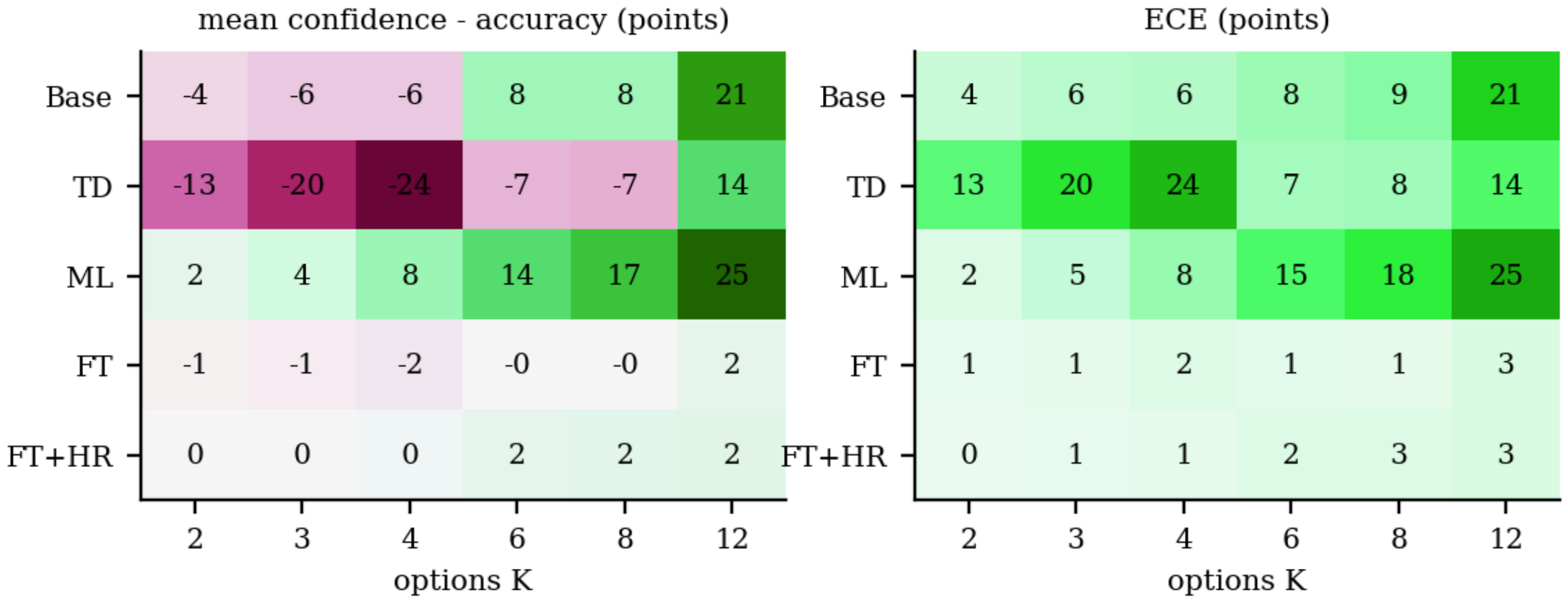

**Figure 22.** Confidence against accuracy by K. Left: over-confidence (red = claims more than it delivers, blue = under-confident). Right: ECE of the top probability on the same draws.

### 14.3 Past twelve options and a second schema

T-06 tests whether the per-doubling slope fitted on K≤12 predicts accuracy at K=24 on the pooled 24-label question. For Base the prediction is 70.2% and the measurement 69.2% (-1.0 points); for FT it is 83.8% against 79.1% (-4.7). The log-linear rule is therefore a usable first-order budget for Base; for FT the measurement is below it by 4.7 points, with an interval that excludes zero. On the HR schema (Table 19, lower block) the fine-tune trained on topics loses 4.8 points per doubling, between its topic-schema slope and Base's 5.5: what the fine-tune learned about K does not fully carry over.

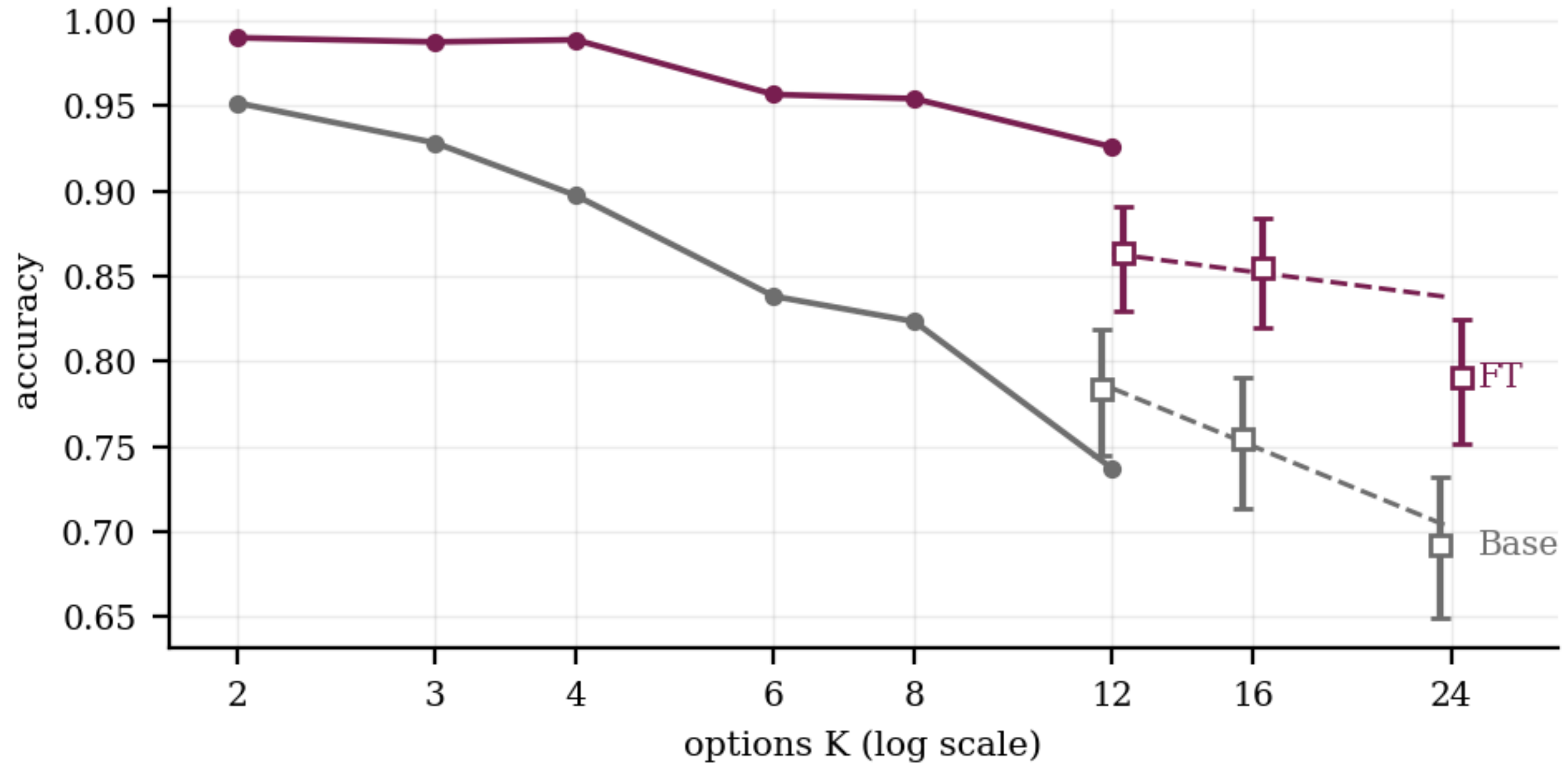

**Figure 23.** Does the log-linear loss continue past K=12? Circles: K-curve on the topic question. Squares: the 24-label pool (95% Wilson intervals). Dashed: prediction for K=24 from the K=12 point of the pool and the slope fitted on K≤12.

## 15 The option-count temperature

### 15.1 What is stored and what is applied (T-03)

The base checkpoint stores 0.1006 for questions with 11 or more options, a divisor that multiplies the logits by 9.9. The runtime does not apply it: `clamp_temperature` confines every temperature to [0.5, 5.0], so the applied value is 0.50 and the logits are still doubled. T-03 fails for Base for that reason (stored and applied differ by a factor of 5.0); the upstream source comment on this clamp says the same thing. The fine-tune's config contains no per-bucket table at all, only the trainer's three per-type values. The choice value, 5.0, is the upper clamp bound, and a fit that sits on a bound may have wanted to go further, so T-03 is inconclusive for FT.

**Table 20.** Temperatures in the released root checkpoint's config, what the runtime applies after clamping to [0.5, 5.0], and the multiplier on the logits (1/T). The last two columns give what the fine-tune's and the multilingual sub-checkpoint's configs apply: the trainer stores one value per question type, and the multilingual config stores 1.0 and no bucket table.

| Bucket | Stored T | Applied T | Logit gain (stored) | Logit gain (applied) | Effect | FT config | ML config |
|---|---|---|---|---|---|---|---|
| `choice:3-5` | 1.7602 | 1.7602 | 0.6× | 0.57× | softens | 5.0 (type value; no table) | 1.0 (all) |
| `choice:6-10` | 1.0000 | 1.0000 | 1.0× | 1.00× | neutral | 5.0 (type value; no table) | 1.0 (all) |
| `score:3-5` | 1.2514 | 1.2514 | 0.8× | 0.80× | softens | 1.0 | 1.0 (all) |
| `noul:2` | 1.9834 | 1.9834 | 0.5× | 0.50× | softens | 1.0 | 1.0 (all) |
| `choice:11+` | 0.1006 | 0.5000 | 9.9× | 2.00× | sharpens | 5.0 (type value; no table) | 1.0 (all) |
| `choice:2` | 1.9064 | 1.9064 | 0.5× | 0.52× | softens | 5.0 (type value; no table) | 1.0 (all) |

### 15.2 Does the applied scale match the data? (T-04)

Re-scaling Base's own probabilities on the 270 twelve-option items shows the direction of the error. Mean top probability is 94.7% as shipped (accuracy 72.2%), 87.8% if the clamp were replaced by neutral T=1, and 99.3% at the stored 0.1006. ECE is 22.5 points as shipped, 16.4 at T=1 and 4.6 at the optimum, which needs a multiplier of 4.3 (total T≈2.2), a *softening* by a factor of 4.3 relative to the runtime. TD wants m≈2.0. After fine-tuning the optimum is m≈1.00, i.e. the trainer's fitted temperature is about right for this question, and T-04 passes for FT with an excess of 0.0 points [0.0-2.3].

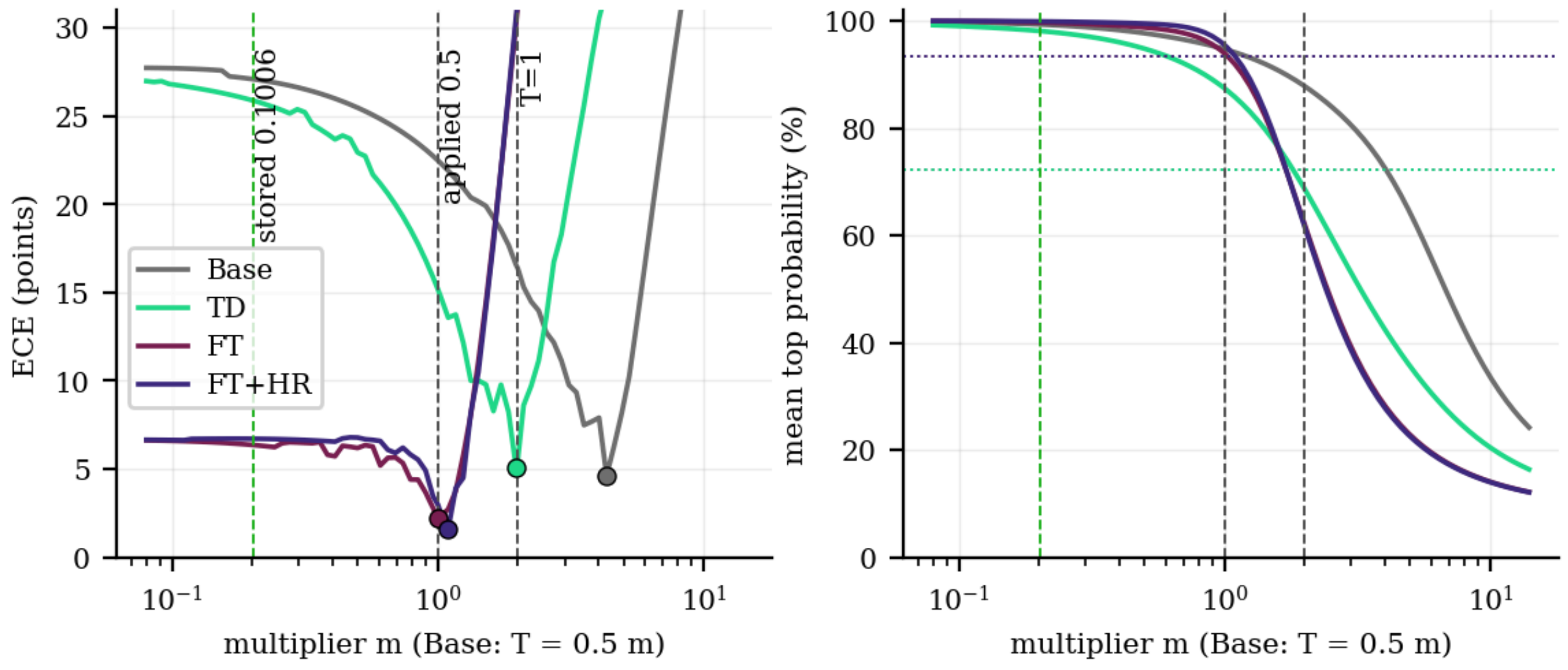

**Figure 24.** Re-scaling the 270 twelve-option items. Left: ECE against the multiplier m applied on top of the runtime's own temperature (dots: minimum). Right: mean top probability, with each model's accuracy as a dotted horizontal line; a model is calibrated where its curve crosses its line. Dashed verticals: the stored 0.1006 (m=0.20), the clamped value actually applied (m=1) and neutral T=1 (m=2) for Base.

**Table 21.** Calibration of the 12-option items at the runtime's own scale (m=1) and at alternative multipliers (270 items). 'best m' minimises ECE; 'NLL m' minimises log loss. The last two ECE columns evaluate m=2 and m=0.2 (for Base: T=1 and the unclamped stored value).

| Model | Acc % | Conf % (m=1) | ECE (m=1) | best m | ECE (best) | NLL m | ECE (m=2) | ECE (m=0.2) | AUROC |
|---|---|---|---|---|---|---|---|---|---|
| **Base** | 72.2 | 94.7 | 22.5 | 4.32 | 4.6 | 4.04 | 16.4 | 27.1 | 0.71 |
| **TD** | 72.2 | 87.4 | 15.2 | 1.97 | 5.0 | 1.73 | 5.5 | 25.9 | 0.71 |
| **FT** | 93.3 | 93.9 | 2.2 | 1.00 | 2.2 | 1.00 | 31.2 | 6.4 | 0.78 |
| **FT+HR** | 93.3 | 95.5 | 3.0 | 1.09 | 1.6 | 1.02 | 31.7 | 6.7 | 0.84 |

### 15.3 Is it only a scale problem?

A temperature cannot reorder items of one question type, so it cannot repair ranking. The AUROC column of Table 21 separates the two cases: for Base the confidence ranks correct above wrong answers with probability 0.71, for FT 0.78. Re-scaling fixes the former model's numbers but not its ability to know which answers are wrong, so a gate on confidence keeps the accuracy of the retained answers only as high as that ranking allows.

## 16 Beyond twelve options

### 16.1 Short descriptions (T-05, T-06)

With label-only or keyword descriptions the pooled question has a head of 185 tokens, so the default budget never cuts it (Section 13). The measurement agrees: raising `head_max_len` to 384 or 768 changes the outcome of 0 of 468 items for Base and 0 for FT, i.e. the token sequences are identical and T-05 passes. Past twelve options the loss is therefore the same K effect as in Section 14 and has nothing to do with the budget: Base goes from 78.4% at K=12 to 69.2% at K=24, FT from 86.3% to 79.1%. Table 22 shows that the K=12 to K=24 step costs the topic items 4.6 points for FT but the HR items 9.2, the schema it had not seen.

**Table 22.** Accuracy (%) on the pooled 24-label question with short descriptions; column headers give K and `head_max_len`. 468 items (195 topic + 273 HR); each item's own label is gold and distractors are drawn from the other 23. The last two columns give the change from K=12 to K=24 (default budget) for the topic and the HR items separately.

| Model | K=12, 192 | K=16, 192 | K=16, 384 | K=24, 192 | K=24, 384 | K=24, 768 | Δ topic (pts) | Δ HR (pts) |
|---|---|---|---|---|---|---|---|---|

| Model | K=12, 192 | K=16, 192 | K=16, 384 | K=24, 192 | K=24, 384 | K=24, 768 | Δ topic (pts) | Δ HR (pts) |
|---|---|---|---|---|---|---|---|---|
| **Base** | 78.4 | 75.4 | 74.1 | 69.2 | 69.2 | 69.2 | -13.8 | -5.9 |
| **TD** | 79.7 | 76.5 | 75.6 | 66.5 | 66.5 | 66.5 | -19.0 | -9.2 |
| **FT** | 86.3 | 85.5 | 84.0 | 79.1 | 79.1 | 79.1 | -4.6 | -9.2 |
| **FT+HR** | 90.6 | 88.7 | 87.6 | 82.3 | 82.3 | 82.3 | -9.7 | -7.3 |

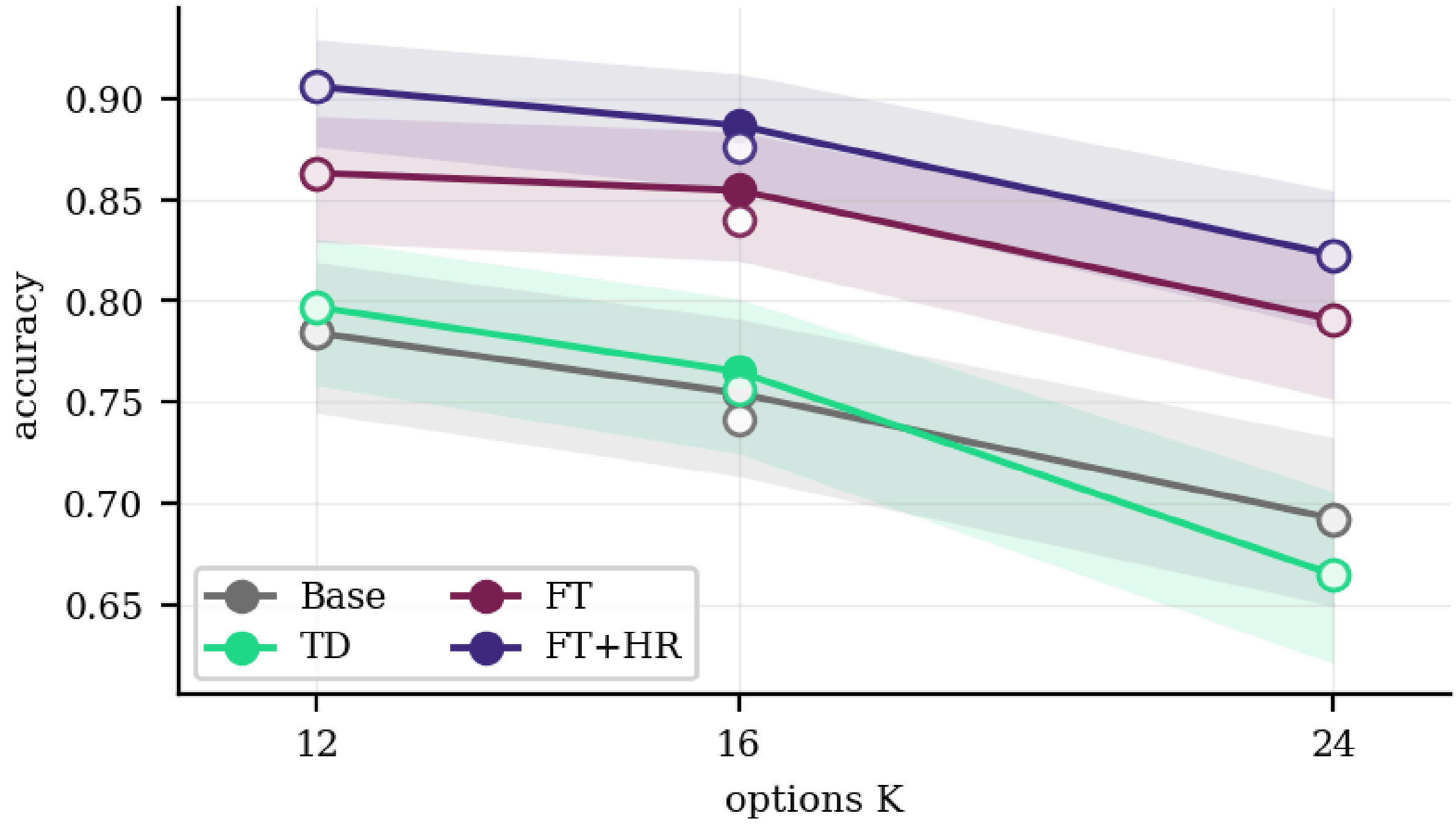

**Figure 25.** Pooled 24-label question, short descriptions. Filled markers and bands: default `head_max_len` 192 (95% Wilson). Hollow markers: `head_max_len` 384 at K=16 and 768 at K=24; they sit exactly on the filled ones.

### 16.2 Long descriptions behind opaque names (T-07, T-08, T-09)

The stress case is built so that the budget matters: names are meaningless (`x0`…`x23`), every description starts with the same preamble and the label-specific words come last. At the default budget all four models are at 2.6-4.9% against a chance of 4.2%, and T-07 fails for both rows of the matrix. The failure has a shape: Base, TD, FT and FT+HR answer at most 0 of the 273 HR items correctly, where a random guesser would get 11.4 on average and none with probability 9.0e-06. The models are not guessing uniformly: whatever they pick, it is never the right option in the second half of the list. At 384 the cap is still inside the preamble and nothing changes. At 768, where 57% of the informative tokens survive, accuracy rises to 36% (TD), 38% (FT) and 56% (FT+HR), but only 17% for Base, which fails T-08. At 1536, where nothing is cut, accuracy falls again for FT (17%) and Base (10%), plausibly because a 772-token head is far outside what any of these models saw in training; we have not tested that explanation.

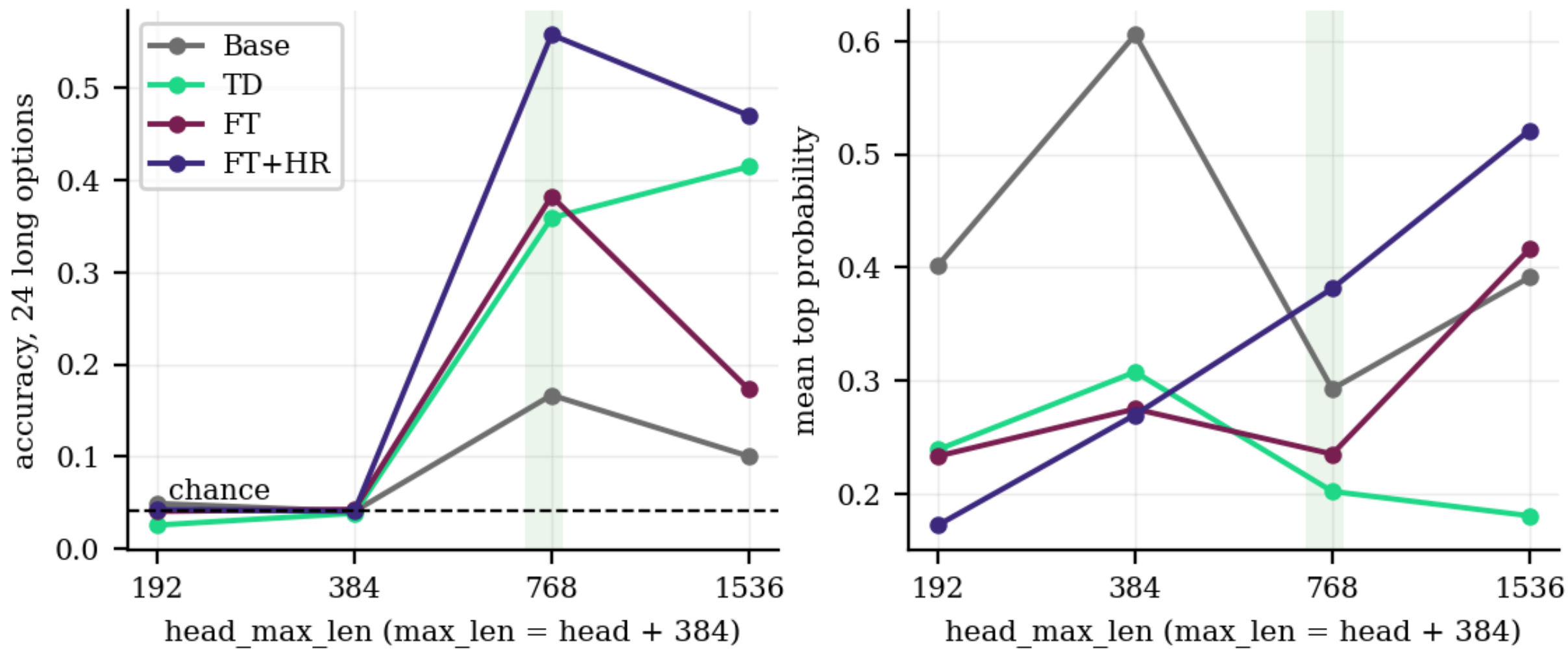

**Figure 26.** Long-description question (24 opaque names). Left: accuracy against `head_max_len` with `max_len` raised to head+384. Right: mean reported confidence in the same runs. The shaded band marks 768, the only setting at which part of the informative text survives the head budget.

For T-09, the runtime reports collapsed options in `usage["options"]` only when two options end up with the same token span. Here every option keeps its distinct name, so the share of flagged questions is 0% at the default budget although no label-specific token survives; T-09 fails. A caller can still detect the situation indirectly (Table 23): mean confidence falls from 83% on the short K=24 question to 23% on the long one for FT, with 94% of questions below 0.5 against 10%. For Base the signal is weaker (mean 40%, 80% below 0.5), consistent with the over-confidence of Section 15.

**Table 23.** Mean top probability (%) in the 24-option runs, and the share of questions whose top probability is below 0.5: a possible detection signal for the silent truncation of T-09.

| Model | short, K=24 | long, 192 | long, 768 | long 192: conf < 0.5 (%) | short K=24: conf < 0.5 (%) |
|---|---|---|---|---|---|
| **Base** | 89 | 40 | 29 | 80 | 7 |
| **TD** | 77 | 24 | 20 | 96 | 14 |
| **FT** | 83 | 23 | 23 | 94 | 10 |
| **FT+HR** | 89 | 17 | 38 | 88 | 1 |

## 17 Order, position and control tests

### 17.1 Order sensitivity (T-11)

The encoder places option *s* after option *s–1*, so an option's representation can depend on where it is listed. For Base, 31% of the 270 items are answered correctly under some orderings and wrongly under others (T-11 fails: [25.9-37.4] against a limit of 15%); for FT the share is 11% and the test passes. The spread by position is also large: Base's gold-slot accuracy ranges from 65% to 81% across positions (Figure 27), a spread of 16 points against 4 for FT. Majority vote over orderings is no cure for either model: it returns 75.2% for Base, against a single-ordering mean of 73.8%, because the always-wrong items (14%) are wrong in every ordering.

**Table 24.** Order sensitivity at K=12: 24 random orderings of the 12 topics for each of 270 items (6,480 forward passes per system). 'Flip' = items answered correctly under some orderings and wrongly under others; 'slot' columns give accuracy by the position of the gold option (worst / best of twelve positions) and their spread in points.

| System | Acc % | always right % | always wrong % | flip % [95% CI] | majority-vote acc % | worst / best slot | spread (pts) |
|---|---|---|---|---|---|---|---|
| **Base** | 73.8 | 54 | 14 | 31.5 [25.9-37.4] | 75.2 | 65 / 81 | 15.9 |

| System | Acc % | always right % | always wrong % | flip % [95% CI] | majority-vote acc % | worst / best slot | spread (pts) |
|---|---|---|---|---|---|---|---|
| **TD** | 74.0 | 61 | 17 | 22.2 [17.4-27.4] | 74.4 | 66 / 82 | 16.2 |
| **FT** | 92.7 | 86 | 3 | 10.7 [7.4-14.4] | 92.6 | 90 / 95 | 4.2 |
| **FT-noshuf** | 92.3 | 85 | 2 | 13.0 [9.6-17.0] | 93.0 | 90 / 96 | 6.6 |
| **FT-par** | 92.2 | 92 | 8 | 0.0 [0.0-0.0] | 92.2 | 88 / 95 | 7.1 |

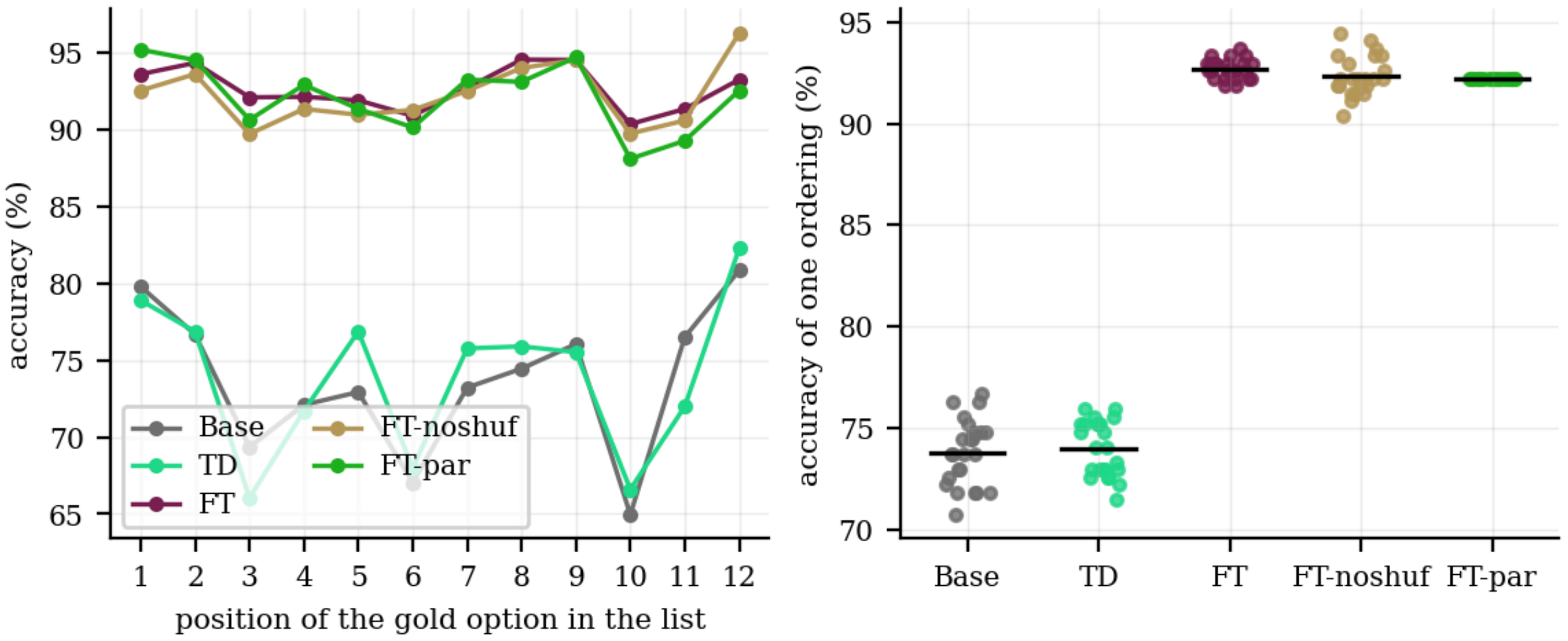

**Figure 27.** Left: accuracy by the position at which the gold option is listed (about 540 passes per position and system). Right: accuracy of each of the 24 orderings taken one at a time (dots) and their mean (bar).

The architectural variant trained with the parallel option layout (FT-par) is the one system with a flip share of 0%: every option starts at the same position and cannot attend to another, so reordering only permutes the markers. That makes it a built-in correctness check for this rig, since a harness bug that mixed up gold positions would show up as flips there; and its accuracy, 92.2%, is within 0.5 points of FT's 92.7%. Its slot spread (7 points) is not smaller than FT's (4), so reordering leaves its answers unchanged but does not equalise accuracy across positions.

## 17.2 Slot preference (T-12) and controls (T-20)

With twelve identical options and unrelated text, an unbiased model would give each slot 1/12 of the probability. Base puts up to 2.17× that on one slot (first slot 1.66×) and the fine-tune 1.54× (Table 25); with only 120 texts the estimate carries no interval, so the verdict for Base is *inconclusive*. The two-, three- and five-option cases are close to uniform for every system (Table 25, K=3 and 5).

**Table 25.** Largest slot probability / first-slot probability, in multiples of uniform, with identical options.

| System | K=3 | K=5 | K=12 |
|---|---|---|---|
| **Base** | 1.18 / 1.18 | 1.13 / 1.13 | 2.17 / 1.66 |
| **TD** | 1.11 / 0.96 | 1.19 / 0.95 | 1.68 / 0.87 |
| **ML** | 1.04 / 0.95 | 1.07 / 0.88 | 1.30 / 1.15 |
| **FT** | 1.18 / 1.18 | 1.24 / 1.10 | 1.54 / 1.44 |
| **FT-noshuf** | 1.08 / 1.08 | 1.31 / 1.05 | 1.19 / 1.19 |
| **FT-par** | 1.00 / 1.00 | 1.04 / 0.97 | 1.11 / 1.04 |

The negative control behaves as it should (T-20): with opaque names and empty descriptions the answer cannot be inferred, and Base (8.5%), TD (8.9%) and FT (9.3%) are at chance. That bears on T-07: accuracy can come only from the text of the option, so the 4% of the long-description runs at the default budget reflects the information that was cut. The remaining rows show how much each system relies on names or on descriptions when only one is present.

**Table 26.** Controls on the 12-topic question (270 items). Chance is 8.3%.

| Condition | Base % | TD % | FT % |
|---|---|---|---|
| no information (opaque names, empty descriptions) | 8.5 [5.7-12.5] | 8.9 [6.0-12.9] | 9.3 [6.4-13.3] |
| names only (descriptions empty) | 78.5 [73.2-83.0] | 76.7 [71.3-81.3] | 91.9 [88.0-94.6] |
| descriptions only (names `x0..x11`) | 70.7 [65.1-75.8] | 70.7 [65.1-75.8] | 91.5 [87.5-94.3] |

## 18 Fine-tuning, transfer and rewording

### 18.1 In-schema effect (T-13)

On the topic schema the fine-tune improves on the released root checkpoint from 73.3% to 92.8% (+19.5 points, p<0.001); T-13 passes. The gain is spread over the topics: the lowest Base recall is 33% for `password`, FT's recall for it is 67%. The most frequent confusions of Base at K=12 (gold to predicted, count; count after fine-tuning) are privacy → security (9, 0 after fine-tuning); invoice → returns (8, 0 after fine-tuning); api → security (8, 1 after fine-tuning). These pairs are semantically adjacent, which suggests that part of the remaining error comes from the label definitions. The three further seeds of the topic fine-tune give a standard deviation of 1.6 points at K=12 (range 89.6-93.0%) and of 0.7 points on the HR items (T-17).

**Table 27.** Full 12-way accuracy on the two schemas with the published option order (topics: 195 items, HR: 273 items). The classical baseline has the same training rows as the fine-tunes for each schema.

| System | Topic acc % [95% CI] | Conf | HR acc % [95% CI] | Conf |
|---|---|---|---|---|
| **Base** | 73.3% [66.7-79.0] | 95% | 82.8% [77.9-86.8] | 90% |
| **TD** | 74.4% [67.8-80.0] | 87% | 82.1% [77.1-86.1] | 82% |
| **ML** | 57.9% [50.9-64.7] | 82% | 50.5% [44.7-56.4] | 79% |
| **FT** | 92.8% [88.3-95.7] | 94% | 84.2% [79.5-88.1] | 84% |
| **FT+HR** | 93.8% [89.6-96.4] | 95% | 90.8% [86.8-93.7] | 94% |
| TF-IDF + LR (trained on the same rows: 1145 topic / 443 HR) | 88.2% [82.9-92.0] | n/a | 90.1% [86.0-93.1] | n/a |

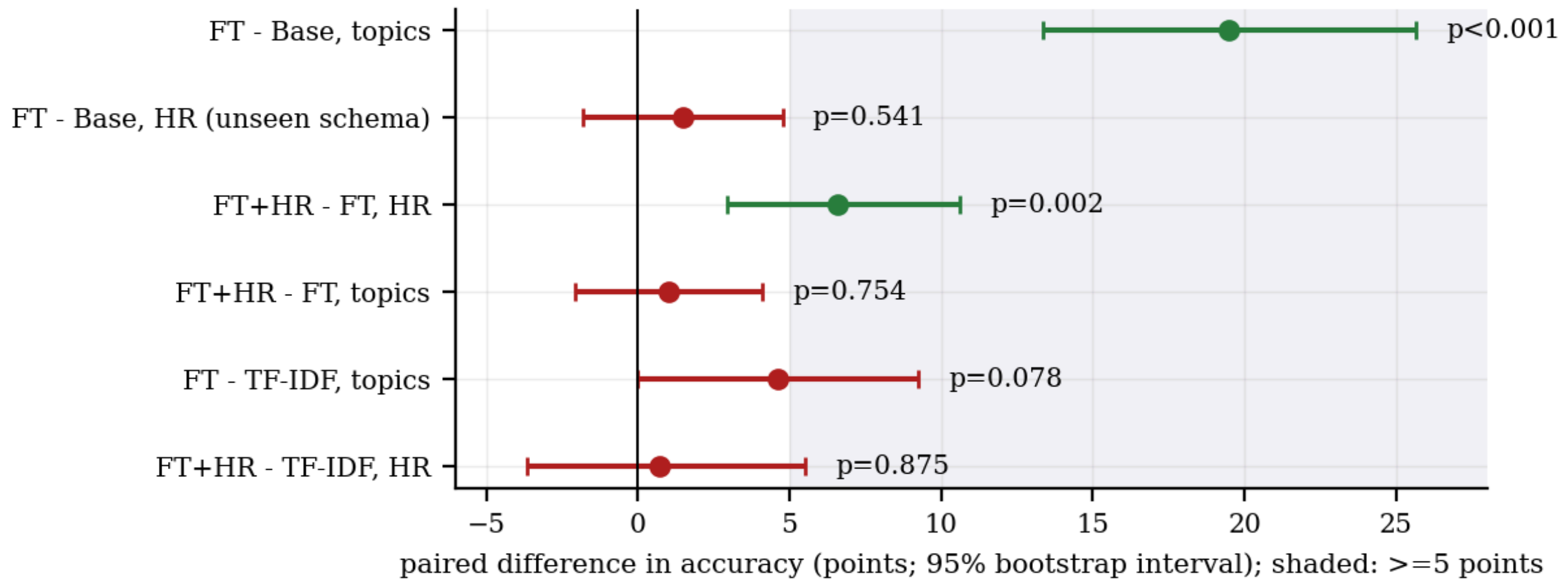

**Figure 28.** Paired accuracy differences on identical items. Green: meets the effect criterion (>=5 points, p<0.01); amber: large or significant but not both; red: below 5 points.

### 18.2 Transfer and adaptation (T-14, T-15, T-16)

A second schema separates a general many-option skill from a topic-specific one. The topic-trained fine-tune moves HR accuracy from 82.8% (best released system, Base) to 84.2%, +1.5 points (p=0.541). That is short of the 5-point bar, so T-14 fails by the stated rule; the 95% interval [-1.8 to 4.8] excludes a gain of the 5-point bar. With a single second schema the data show a small transfer at most. The HR block of Table 19 shows the same small shift at every K. Adding 443 HR rows lifts accuracy to 90.8% (+6.6 points over FT, p=0.002; T-15 passes), more than the spread between seeds of the topic fine-tune on the same items (0.7 points). The topic schema moves by +1.0 points ([-2.1 to 4.1]); the interval is wider than the ±2 points of T-16, so that verdict is inconclusive.

### 18.3 Against a bag-of-n-grams classifier (T-18)

T-18 compares the fine-tunes with TF-IDF word and character n-grams plus logistic regression trained on the same rows. On the topic schema FT is +4.6 points from the classifier (p=0.078); on the HR schema FT+HR is +0.7 (p=0.875). Neither clears the five-point criterion, so T-18 fails: on a single 12-way schema the classical model is a strong control (Figure 29). The encoder's other claimed advantage, answering questions it was not trained on, is measured here only through the HR transfer result.

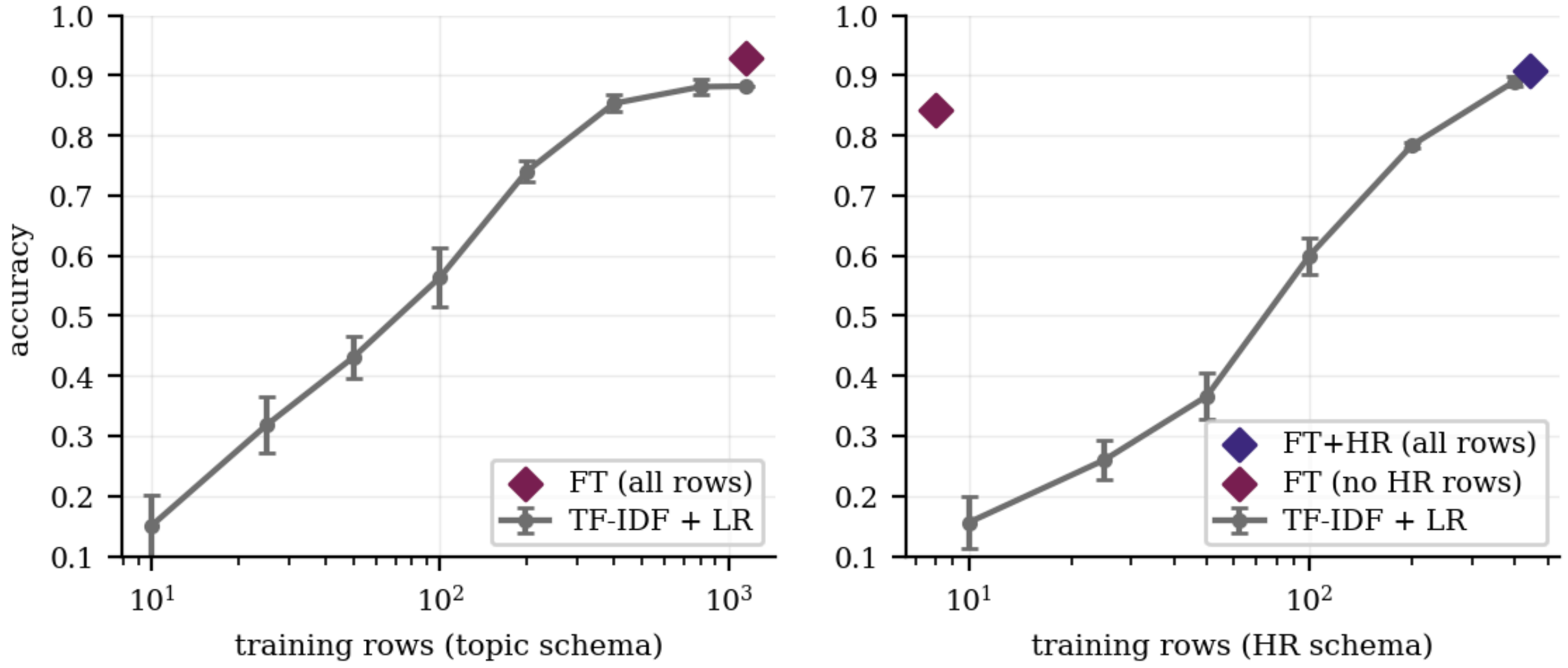

**Figure 29.** Data efficiency: TF-IDF + logistic regression as a function of the number of training rows (five subsamples; bars are one standard deviation) against the fine-tuned encoder at its full training set.

### 18.4 Is the fine-tune brittle to rewording?

The tests above use the text as the teacher wrote it. Table 28 re-runs the K=12 topic question on mechanically altered texts. FT loses 16 points under `upper` (77% from 93%) and 15 under 15% character noise, against 23 for Base; the TF-IDF baseline loses 1 under the same noise. So under character noise the fine-tune loses no more than the released model in absolute points. Its loss under capitalisation and truncation suggests that part of its clean accuracy is specific to the teacher's register. This qualifies every accuracy figure in this report and carries no verdict.

**Table 28.** Topic-schema accuracy (%, 195 items, full 12-way) when the text is reworded mechanically. Parentheses: change against the unchanged text.

| Perturbation | Meaning | Base | TD | FT | TF-IDF |
|---|---|---|---|---|---|
| clean | unchanged | 73.3 | 74.4 | 92.8 | 88.2 |
| typo5 | 5% of characters edited | 64.6 (-9) | 66.7 (-8) | 88.7 (-4) | 87.2 (-1) |
| typo15 | 15% of characters edited | 50.3 (-23) | 56.4 (-18) | 77.4 (-15) | 87.7 (-1) |
| upper | all capitals | 47.2 (-26) | 43.6 (-31) | 76.9 (-16) | 88.2 (+0) |
| lower_nopunct | lower case, punctuation removed | 73.8 (+1) | 70.3 (-4) | 90.8 (-2) | 89.2 (+1) |
| first_half | only the first half of the words | 59.5 (-14) | 59.0 (-15) | 81.5 (-11) | 85.6 (-3) |
| strip_tone | lower case, exclamation marks removed | 73.8 (+1) | 75.9 (+2) | 91.8 (-1) | 88.2 (+0) |

## 19 Validity of the harness

### 19.1 Label provenance

Every label in this report comes from Qwen3.8-27B. For the topic schema the teacher wrote 240 candidates and 195 survived the blind re-labelling pass (81%); for the HR schema 716 of 773 did (93%), which are split into 443 training and 273 test rows. The filter shares the teacher's family, so by itself it says little about label quality. T-19 therefore adds a model from a different family (gemma4-31b) that labels every held-out item blind: it agrees with the Qwen label on 94.4% of the topic items [90.2-96.8] and on 96.7% of the HR items [93.9-98.3], and the test passes. Residual disagreement is the label noise that two strong models do not share, and amounts to 11 and 9 items.

**Table 29.** Accuracy (%) on all items, on the items where the independent LLM agrees with the Qwen label, and on the items where it does not (topics: 11 of 195 disagree; HR: 9 of 273).

| Schema | System | All items | LLM agrees | LLM disagrees | Δ agreed vs all (pts) |
|---|---|---|---|---|---|
| topics | **Base** | 73.3 | 77.2 | 9.1 | +3.8 |
| topics | **TD** | 74.4 | 77.7 | 18.2 | +3.4 |
| topics | **ML** | 57.9 | 60.3 | 18.2 | +2.4 |
| topics | **FT** | 92.8 | 94.6 | 63.6 | +1.7 |
| topics | **FT+HR** | 93.8 | 95.7 | 63.6 | +1.8 |
| HR | **Base** | 82.8 | 83.7 | 55.6 | +0.9 |
| HR | **TD** | 82.1 | 83.0 | 55.6 | +0.9 |
| HR | **ML** | 50.5 | 51.5 | 22.2 | +1.0 |
| HR | **FT** | 84.2 | 85.6 | 44.4 | +1.4 |
| HR | **FT+HR** | 90.8 | 92.0 | 55.6 | +1.2 |

### 19.2 Does label noise explain a verdict?

Restricting to the agreed items raises every system's accuracy by a similar amount and leaves the ordering of systems unchanged. The headline effects survive: FT+HR over FT on HR items is +6.4 points on the agreed subset (p=0.003), and the topic schema changes by +1.1 (p=0.727). The most error-prone items are therefore partly label noise, but not the reason for any pass or fail in the matrix. What this check cannot do is detect errors on which both LLMs agree; a human-labelled replication of the two 12-way sets remains the missing control.

## 20 Failure-mode catalogue

Ten failure modes came out of the twenty tests. Only one of them (FM-10, an undersized `max_len`) stops the request; the other nine return an answer with a confidence attached.

**Table 30.** Failure-mode catalogue. Magnitudes are Base / FT where both are defined; 'visible' says whether the runtime tells the caller. FM-4 and FM-10 are read from the runtime code and checked with the runtime's own functions, not by model runs.

| ID | Mode | Trigger | Magnitude | Visible? | Mitigation |
|---|---|---|---|---|---|
| FM-1 | Accuracy erodes with K | adding distractors | 21.5 / 6.4 pts (K=2→12) | no | fine-tune on the schema (T-01 passes); keep K small |
| FM-2 | Confidence rises while accuracy falls | K ≥ 11 (bucket 11+, stored T=0.1006 clamped to 0.5) | 21.5 / 1.7 pts | no | refit the temperature on held-out items; Base best total T≈2.2 |

| ID | Mode | Trigger | Magnitude | Visible? | Mitigation |
|---|---|---|---|---|---|
| FM-3 | Description text silently cut | K×(option length) > `head_max_len`−16 | 11.8 / 34.2 pts regained by 768 | no (flag needs identical option spans) | measure the head with `build_head`; raise `head_max_len` and `max_len` together |
| FM-4 | Text dropped by an over-large head | head ≥ `max_len`−1 with all markers inside | text room 0 tokens at (1024, 768), static | partly (`usage.truncated`) | keep `max_len` ≥ head + expected text + 1 |
| FM-5 | Answer depends on list order | K=12, sequential layout | 15.9 / 4.2 pts slot spread; flips 31% / 11% | no | shuffle options in training; parallel layout gives zero flips |
| FM-6 | Small transfer to a new schema | new schema with the same K | transfer +1.5 vs in-schema +19.5 | no | add rows of the target schema (443 rows: +6.6 pts) |
| FM-7 | Too-wide head hurts | `head_max_len` 1536 (head 772 tokens) | 6.6 / 20.9 pts lost vs 768 | no | do not go beyond the width the model was trained for; test the canary |
| FM-8 | Adjacent topics confused | overlapping labels (login/password, invoice/refund) | 36.5 / 24.5 pts (worst topic) | no | merge or redefine labels; human review of the confused pairs |
| FM-9 | Slot preference with identical options | K=12, uninformative text | max slot 2.17× / 1.54× uniform | no | none needed for K≤5; monitor at K=12 |
| FM-10 | Too-small `max_len` | head > `max_len` and markers cut | 17 of 24 markers (static) | **yes**: `ValueError` | raise `max_len` or lower `head_max_len` |

The catalogue is ordered by when a deployment would meet each mode. FM-1 and FM-2 appear once a question has more than about eight options. FM-3, FM-4 and FM-7 need descriptions long enough to hit the head budget, and FM-5 and FM-9 need K=12. FM-6 appears when a model fitted on one schema meets another, and FM-8 is a label-design issue that no setting removes. The largest single loss is the 34-point gap between the default and the raised head budget for long descriptions, and the runtime is best placed to detect that one.

## 21 Recommendations

**Table 31.** Configuration advice. Each row names the evidence in this report and a check that can be automated.

| ID | Advice | Evidence | Check |
|---|---|---|---|
| CA-1 | Do not read the reported confidence of a question with 11+ options as a probability. Refit a multiplier on held-out items of your own schema before gating on it. | T-02, T-04: Base over-states by 21 points at K=12; best total T is about 2.2, not 0.5 | ECE and a reliability diagram on ≥200 held-out items |
| CA-2 | Before deployment compute the head with `build_head`; require `tokens_per_option` to be `None` (no cut) for every question. | T-05, Section 13: a 185-token head fits 192; a 181-token head built from long options is already cut to 7 tokens per option | static check in CI; no model needed |
| CA-3 | If descriptions carry the information, raise `head_max_len` and `max_len` together (head + room for the longest text) and re-measure; do not stop at `max_len=512`. | T-07/T-08/T-10: 4% at 192, 38% at 768 for FT; the runtime refuses 768 under 512 | canary question with a known answer per option |
| CA-4 | Put the words that distinguish an option first, and keep descriptions short; both reduce what the head budget can cut. | Section 13: cuts remove the end of an option | decode the head and read it |

| ID | Advice | Evidence | Check |
|---|---|---|---|
| CA-5 | Budget K with the fitted law, and treat anything above about 16 options as a different design problem. A shortlist built from the encoder's own scores did not help in our tests. | T-06 and Section 5: roughly 8 points per doubling for Base; shortlist gains not distinguishable from zero (Section 8) | measure K at your operating point; check a held-out K |
| CA-6 | Fine-tune on rows of the schema you deploy; expect at most a small gain from another schema. | T-14, T-15: +1.5 points transferred, +6.6 from 443 in-schema rows | paired test on identical held-out items |
| CA-7 | If answers must not depend on option order, use the parallel layout or shuffle during training and test with several orderings. | T-11: flip share 31% (Base), 11% (FT), 0% (FT-par) | 24 random orderings per item |
| CA-8 | Keep a TF-IDF + logistic-regression model on the same rows as a control; for a single 12-way schema it is within a few points. | T-18: +4.6 (topics), +0.7 (HR) points | paired McNemar against the encoder |
| CA-9 | Do not compare a probability across option counts or across lists. The gap between the gold option and a fixed rival changes with K in both models; recompute any threshold when the list changes. | Section 6: the raw gap moves by 0.24 to 2.07 nats per doubling of K | measure the gap at your K values |
| CA-10 | Calibrate confidence and conformal sets on items of the schema you deploy; a threshold from another schema over- or under-covers. | Section 8: coverage of a 90% set calibrated on topics is 84% on HR for FT and 99% (sets of 5.9 options) for Base | coverage on held-out items of the target schema |
| CA-11 | Shuffle the option order per request where the application allows it; an early-slot advantage appears on schemas the model did not train on. | Section 7: exploratory trend of 6 to 8 points from first to last slot | accuracy by gold slot on a held-out set |

## 22 Limitations

- Two models of one family. Base is the released root checkpoint and FT is our fine-tune of it; the exponent c, the sign of the IIA shift and the failure of every fix are results about this encoder. Another typed-decision encoder may sit elsewhere.
- Labels are model-made and all text is synthetic English. Qwen3.8-27B wrote and labelled the items (Section 19 gives the agreement with a second model), so label noise that both models share is invisible. Test texts are short, so the text is never truncated and only the option block is cut by the head budget.
- The law is a fit, not a derivation. It has two parameters fitted on four to six points per cell, and the interval of c is wide in several cells (Table 6). The band of 0.68 to 0.88 describes four pools built from five schemas of one family, and the Mega-35 failure shows that a pool of unrelated schemas with confusable labels breaks it. We did not test labels that were chosen to be hard.
- Distractors are sampled uniformly from the pool. Real label sets contain families of near-synonyms, and we expect harder rivals there, but we did not run a set built that way.
- The IIA test fixes the rival at the first distractor of a draw and not at the strongest wrong label, and stored probabilities were rounded to four decimals, with a floor of 1e-4. Base stops at K=10 for that reason. The temperature division by the runtime was read from the code and applied to every probability; a different post-processing would change the absolute slopes and not their sign.
- The position test has about 80 to 250 cases per slot, so slot effects below about two points cannot be excluded. The trend test was run after the pre-stated spread test and is exploratory.
- The fixes are tested on ten orderings at most, on half of one schema for the conformal calibration, and with the shortlist chosen on the test items. Per-option scoring, the first stage of Runoff, was not run, and neither was an order-invariant model.
- Runs are frozen, with one training run for FT in the new experiments. Intervals cover item sampling and not training randomness; T-17 of the first part covers four seeds.

- Thresholds of the twenty tests of Part II are conventions, set when the test plan was written and applied identically to every system. Table 17 shows the range over which each verdict holds. Static tests (T-03, T-10) are code readings and were not exercised through a forward pass. Fine-tunes are fragile to rewording (Section 18.4), so accuracy on teacher-style text overstates what a noisy deployment would see.

## 23 Conclusion

Accuracy of Laya against the number of options K follows logit(accuracy) = a - c log(K-1) with c between 0.68 and 0.88 in all eight cells (two models, four pools). Fitted on K up to 6, it predicts K from 7 to 12 within 0.6 points on average, and K up to 24 within 1.2. With the slope borrowed from the topic schema and only the two-option accuracy measured, it predicts every other K on a schema that the fine-tune never saw within 0.3 (FT) and 0.7 (Base) points. It fails on a 35-label pool of five schemas when it is fitted on two points of that pool. IIA is rejected for both models: adding options raises the gold-versus-rival gap by +0.24 to +2.07 nats per doubling of K, and the pairwise answers still predict the K-way accuracy to within 4.7 points while missing its probability by 9 to 23 points. The strongest rival, not the number of rivals, explains the errors.

Most of the fixes taken from the decoder literature did not transfer. Slot has little effect, averaging over ten orderings gains at most 2.2 points with intervals that include zero, a shortlist and a temperature law fitted on one schema do not help on another, and a conformal set calibrated on the topic schema covers 84% on HR-12 for FT when it should cover 90%. 2 of the ten predictions that we wrote before the runs held, one of them only up to K=24, and 8 were refuted in full or in part. Part II keeps the original twenty tests: the released checkpoint passes 5 of the 14 that apply to it and the fine-tune 13 of 20, each against a threshold that remains a convention. For a user the practical rule is short: measure two-option accuracy on your own schema, apply an exponent of about 0.75 to budget K, check the rule on a held-out K when the label set mixes domains, and do not read a probability without the list it was computed over.

# Appendix A Test run sheets

**Table 32.** Frozen inputs of each test (first 16 hex digits of the SHA-256 of the file).

| File | Used by | SHA-256 prefix |
|---|---|---|
| `results/e0__v1.json` | T-13, T-16, T-19 (topic items) | `8ce8364ad4cb7514` |
| `results/e3__base.json` | T-01, T-02, T-17 (K-curves; also td, mlo, v1, v1_hr, v1_s1..s3) | `286dd24e1a767207` |
| `results/e10__v1_hr.json` | T-14, T-15, T-16 (HR schema; also every system) | `213b5fba342bd21f` |
| `results/e11__base.json` | T-20 | `2e357cbd40f6b03f` |
| `results/e12__base.json` | T-12 | `b3e7c80198b4325e` |
| `results/e13__base.json` | T-11 (also td, v1, v1_noshuf, v1_parallel) | `c64df0393c5681e6` |
| `results/e14__v1_hr.json` | T-05..T-09 (also base, td, v1) | `e144533f8c0b0ebc` |
| `results/e15__v1.json` | Section 18.4 | `32879cb6cca8e946` |
| `results/xlabel_gemma4-31b.json` | T-19 | `28f1fa32695a39f6` |
| `results/baselines.json` | T-18, Section 18.4 | `2cd5bf98e456091f` |
| `data/many_test.jsonl` | topic test set | `9897ea728a2dd620` |
| `data/hr_test.jsonl` | HR test set | `d819fc43ebcafc58` |
| `data/hr_train.jsonl` | HR training rows | `fa6bff19c4d43c7b` |
| `results/a6_fix__hr__base.json` | shortlist runs (8) | `c2c098d0da7452ac` |
| `results/a6_fix__hr__v1.json` | shortlist runs (8) | `25c06ebaf6405cc8` |
| `results/a6_fix__pool24__base.json` | shortlist runs (8) | `4ce58e401af05826` |
| `results/a6_fix__pool24__v1.json` | shortlist runs (8) | `f9f6e4178973cf5b` |
| `results/a6_fix__topic__base.json` | shortlist runs (8) | `969ee8053ef5d6e9` |
| `results/a6_fix__topic__v1.json` | shortlist runs (8) | `97bd80b982e43859` |
| `results/a6_kc__hr__base.json` | nested K-curves (Sections 5, 6, 7, 8, 9) | `87e5045d2ed25631` |
| `results/a6_kc__hr__v1.json` | nested K-curves (Sections 5, 6, 7, 8, 9) | `8e83b39b28c28fd0` |
| `results/a6_kc__mega__base.json` | nested K-curves (Sections 5, 6, 7, 8, 9) | `6418aacaf51a5355` |
| `results/a6_kc__mega__v1.json` | nested K-curves (Sections 5, 6, 7, 8, 9) | `98b629da05ffda5e` |
| `results/a6_kc__megaH__base.json` | nested K-curves (Sections 5, 6, 7, 8, 9) | `34ecfabe57a240cb` |
| `results/a6_kc__megaH__v1.json` | nested K-curves (Sections 5, 6, 7, 8, 9) | `90f17219c4bf4470` |
| `results/a6_kc__pool24__base.json` | nested K-curves (Sections 5, 6, 7, 8, 9) | `5759e62e56cbfd4e` |
| `results/a6_kc__pool24__v1.json` | nested K-curves (Sections 5, 6, 7, 8, 9) | `94cabbe95ffd1b51` |
| `results/a6_kc__topic__base.json` | nested K-curves (Sections 5, 6, 7, 8, 9) | `b8beecd195c38f2b` |
| `results/a6_kc__topic__v1.json` | nested K-curves (Sections 5, 6, 7, 8, 9) | `3ed8cd072558f39f` |
| `results/a6_pairs__hr__base.json` | all-pairs two-option scores (6) | `d29742c366c66e5e` |
| `results/a6_pairs__hr__v1.json` | all-pairs two-option scores (6) | `539ac315141f1d6a` |
| `results/a6_pairs__mega__base.json` | all-pairs two-option scores (6) | `622919209a7d0572` |
| `results/a6_pairs__mega__v1.json` | all-pairs two-option scores (6) | `509673541b96d028` |
| `results/a6_pairs__pool24__base.json` | all-pairs two-option scores (6) | `687dbda18b87205c` |
| `results/a6_pairs__pool24__v1.json` | all-pairs two-option scores (6) | `25cbca6a538218b7` |
| `results/a6_pairs__topic__base.json` | all-pairs two-option scores (6) | `50501d28f043e52c` |

| File | Used by | SHA-256 prefix |
|---|---|---|
| `results/a6_pairs__topic__v1.json` | all-pairs two-option scores (6) | `0f8178ea20031997` |
| `models/v1/model.safetensors` | FT weights | `632868e784f96b31` |
| `models/v1_hr/model.safetensors` | FT+HR weights | `3d8fcd4c3c59ca48` |

# Appendix B  Worked examples

## B.1  Confident errors at K=12

**Table 33.** The five most confident wrong answers of the released root checkpoint at K=12 (all twelve topics listed in their published order), with the fine-tune's answer to the same item.

| Text (first 170 characters) | Gold | Base answer (conf) | FT answer (conf) |
|---|---|---|---|
| Just need to swap out my current login token for a new one. I am planning to change it for security reasons after leaving my last job. | `password` | `login` (1.00) | `password` (0.87) |
| The login page keeps telling me my credentials are incorrect, even though I am double-checking what I am typing. I think I might have changed it on my other laptop and am... | `password` | `login` (1.00) | `login` (0.98) |
| Who else has seen my bank transaction history? I see a log entry indicating a data export was performed on March 12th, but I never authorized a download. I am concerned a... | `privacy` | `security` (1.00) | `privacy` (0.93) |
| Please generate an RMA number so I can process the refund for the unsatisfactory software license key. | `returns` | `refund` (1.00) | `refund` (0.97) |
| Can I add an extra gift wrap option? The label hasn't been generated yet, so I hope there's still time to make changes. | `shipping` | `feature_request` (1.00) | `feature_request` (0.96) |

## B.2  What survives the head budget

**Table 34.** What the encoder reads for two of the 24 long options at each head budget: token strings of the option span as produced by the runtime (the first token is the `[MASK]` marker).

| head_max_len | Option | Surviving tokens |
|---|---|---|
| 192 | x0 (topic) | `[MASK] ∣ ·x ∣ 0 ∣ : ∣ ·a ∣ ·generic ∣ ·category` |
| 192 | x12 (HR) | `[MASK] ∣ ·x ∣ 12 ∣ : ∣ ·a ∣ ·generic ∣ ·category` |
| 768 | x0 (topic) | `[MASK] ∣ ·x ∣ 0 ∣ : ∣ ·a ∣ ·generic ∣ ·category ∣ ·in ∣ ·our ∣ ·routing ∣ ·tax ∣ onomy ∣ , ∣ ·one ∣ ·of ∣ ·twenty ∣ - ∣ four ∣ , ∣ ·used ∣ ·to ∣ ·sort ∣ ·incoming ∣ ·messages ∣ ; ∣ ·typical ∣ ·content ∣ : ∣ ·login` |
| 768 | x12 (HR) | `[MASK] ∣ ·x ∣ 12 ∣ : ∣ ·a ∣ ·generic ∣ ·category ∣ ·in ∣ ·our ∣ ·routing ∣ ·tax ∣ onomy ∣ , ∣ ·one ∣ ·of ∣ ·twenty ∣ - ∣ four ∣ , ∣ ·used ∣ ·to ∣ ·sort ∣ ·incoming ∣ ·messages ∣ ; ∣ ·typical ∣ ·content ∣ : ∣ ·salary ∣ , ∣ ·pay` |

# Appendix C  Full scaling table

**Table 35.** Topic schema: accuracy (%) with 95% bootstrap intervals (over 270 items) for every K.

| System | K=2 | K=3 | K=4 | K=6 | K=8 | K=12 |
|---|---|---|---|---|---|---|
| **Base** | 95.2 [93-97] | 92.8 [91-95] | 89.8 [87-92] | 83.8 [81-87] | 82.3 [79-86] | 73.7 [69-78] |
| **TD** | 94.2 [92-96] | 92.6 [90-95] | 89.8 [87-92] | 84.2 [81-87] | 80.4 [76-84] | 74.2 [69-79] |
| **ML** | 92.0 [89-94] | 85.9 [82-89] | 79.3 [75-83] | 71.4 [67-75] | 66.5 [62-71] | 56.8 [52-62] |
| **FT** | 99.0 [98-100] | 98.8 [98-100] | 98.9 [98-100] | 95.7 [94-97] | 95.4 [93-97] | 92.6 [90-96] |

| System | K=2 | K=3 | K=4 | K=6 | K=8 | K=12 |
|---|---|---|---|---|---|---|
| **FT+HR** | 99.0 [98-100] | 98.3 [97-99] | 97.7 [96-99] | 95.7 [94-97] | 94.9 [93-97] | 93.0 [90-96] |
| **FT seed 1** | 99.0 [98-100] | 98.3 [97-99] | 98.0 [97-99] | 95.9 [94-98] | 95.3 [93-97] | 93.0 [90-96] |
| **FT seed 2** | 98.9 [98-100] | 98.4 [97-99] | 98.0 [97-99] | 95.2 [93-97] | 94.6 [92-97] | 92.8 [90-96] |
| **FT seed 3** | 98.6 [98-100] | 97.8 [97-99] | 97.7 [96-99] | 93.7 [91-96] | 93.0 [90-95] | 89.6 [86-93] |

## Appendix D Environment and rerun

Python 3.10.9, torch 2.11.0+cu128, transformers 5.20.2, laya 0.4.26, numpy 2.2.6. All forward passes ran on one RTX 4090 (24 GB) in bf16 with batch size 16 (a K=12 pass costs about 1.6 ms per item). The frozen runs are produced by `python lab.py <experiment> <system>` with the experiments listed in Table 32 (K-curves: e3 and e10; controls: e11; position prior: e12; 24-order replay: e13; beyond twelve options: e14; rewording: e15). This document recomputes every number from the stored per-item arrays, and the static budget analysis from the tokenizer of the released checkpoint, so rebuilding it requires no GPU. The new experiments of Sections 5 to 9 come from `python a6_kc.py <model> <pool>` (nested K-curves), `python a6_pairs.py <model>` (all two-option questions) and `python a6_fix.py <model>` (shortlists); each writes one resumable file per model and pool to `results/a6_*.json`, and `a6_ana.py` and `a6_new.py` compute every number from them. The text was drafted and edited with LLM assistance; all numbers are recomputed from the stored results by the build scripts and were checked by the authors.